%% file: main.tex
\documentclass[10pt,twocolumn,letterpaper]{article}

\usepackage[pagenumbers]{cvpr} % To force page numbers, e.g. for an arXiv version

\usepackage{multicol}
\usepackage[toc,page]{appendix}
\usepackage[normalem]{ulem}


\definecolor{cvprblue}{rgb}{0.21,0.49,0.74}
\usepackage[pagebackref,breaklinks,colorlinks,allcolors=cvprblue]{hyperref}

\def\paperID{11433} % *** Enter the Paper ID here
\def\confName{CVPR}
\def\confYear{2025}

\title{Temporally Propagated Masks and Bounding Boxes:\\Combining the Best of Both Worlds for Multi-Object Tracking}

\author{Tomasz Stanczyk\\
Inria centre at Université Côte d’Azur\\
2004 Rte des Lucioles, 06902 Valbonne, France\\
{\tt\small tomasz.stanczyk@inria.fr}
\and
Francois Bremond\\
Inria centre at Université Côte d’Azur\\
2004 Rte des Lucioles, 06902 Valbonne, France\\
{\tt\small francois.bremond@inria.fr}
}

\begin{document}
\maketitle
% \input{sec/0_abstract}    
% \input{sec/1_intro}
% \input{sec/2_formatting}
% \input{sec/3_finalcopy}

%%%%%%%%% ABSTRACT
\begin{abstract}
   Multi-object tracking (MOT) involves identifying and consistently tracking objects across video sequences. Traditional tracking-by-detection methods, while effective, often require extensive tuning and lack generalizability. On the other hand, segmentation mask-based methods are more generic but struggle with tracking management, making them unsuitable for MOT. We propose a novel approach, McByte, which incorporates a temporally propagated segmentation mask as a strong association cue within a tracking-by-detection framework. By combining bounding box and propagated mask information, McByte enhances robustness and generalizability without per-sequence tuning. Evaluated on four benchmark datasets - DanceTrack, MOT17, SoccerNet-tracking 2022, and KITTI-tracking - McByte demonstrates performance gain in all cases examined. At the same time, it outperforms existing mask-based methods. Implementation code will be provided upon acceptance.
\end{abstract}

%%% --- THAT'S THE GOOD ONE (Second visual exemple)! --- %%%
% \begin{figure*}
% \centering
% \begin{tabular}{cccc}
% \includegraphics[height=4.0cm]{images/bt2a.png}&
% \includegraphics[height=4.0cm]{images/bt2b.png}&
% \includegraphics[height=4.0cm]{images/ours2a.png}&
% \includegraphics[height=4.0cm]{images/ours2b.png}
% \\
% Frame 459 (baseline)&Frame 475 (baseline)&Frame 459 (McByte)&Frame 475 (McByte)
% \end{tabular}
% \vspace*{0.3cm}
% \caption{Visual output comparison between the baseline and McByte. With the temporally propagated mask guidance, McByte can handle the association of an ambiguous set of bounding boxes and reduce the identity switches - see the subjects with IDs 59 and 63 on the output of McByte. Input image data from~\cite{mot17_ref}. Best seen in color.}
% \label{fig:visual_differences_2}
% \end{figure*}
%%% --- </> --- %%%

%%%%%%%%% BODY TEXT
\section{Introduction}
\label{sec:intro}

Multi-object tracking (MOT) is a computer vision task that involves tracking objects (e.g., people) across video frames while maintaining consistent object IDs. MOT detects objects in each frame and associates them across consecutive frames. Applications include surveillance, automated behavior analysis (e.g., in hospitals), and autonomous driving, making reliable MOT trackers essential.

Tracking-by-detection methods~\cite{bt_ref,ocsort_ref, deepocsort_ref,cbiou_ref,strongsort_ref,hybridsort_ref} use bounding boxes to detect objects in each frame and associate them with those from previous frames, based on cues like position, appearance, and motion. The resulting matches form "tracklets" over consecutive frames. However, these methods often require extensive hyper-parameter tuning for each dataset or even per single sequence, reducing their generalizability and limiting their application across different datasets.

Segmentation mask-based methods~\cite{deva_ref,sam2_ref} generate masks to cover objects and track them across video frames. Trained on large datasets, these methods aim to capture the semantics of image patches, making them more generic. However, they are not designed for MOT, lacking robust management for tracking multiple entities and struggling to detect new objects entering the scene. Additionally, they rely entirely on mask predictions for object positioning, which can be problematic when the predictions are noisy or inaccurate.

In this paper, we explore using a temporally propagated segmentation mask as an association cue to assess its effectiveness in MOT. We propose a novel tracking-by-detection method that combines mask propagation and bounding boxes to improve the association between tracklets and detections. 
%
% The mask propagation is managed according to the tracklet lifespan, with usage constraints to enhance tracking performance. 
%
The mask propagation is managed according to the tracklet lifespan, while the mask is used in a controlled manner to enhance tracking performance. 
Since the temporal mask propagation model is trained on a large dataset, it makes the entire tracking process more generic. Unlike existing tracking-by-detection methods, our approach does not require tuning hyperparameters for each video sequence.

\begin{figure}
\centering
\begin{tabular}{ccc}
\includegraphics[height=2.9cm]{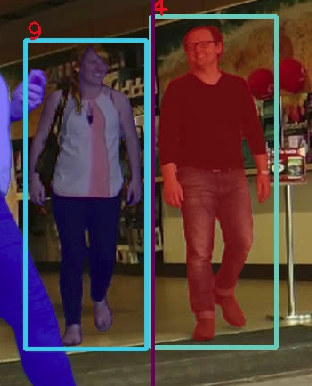}&
\includegraphics[height=2.9cm]{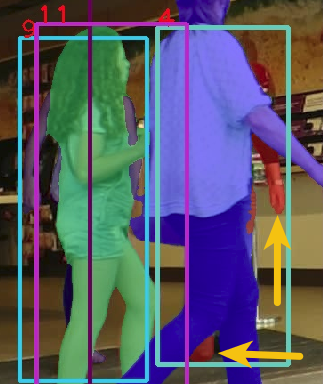}& % &
\includegraphics[height=2.9cm]{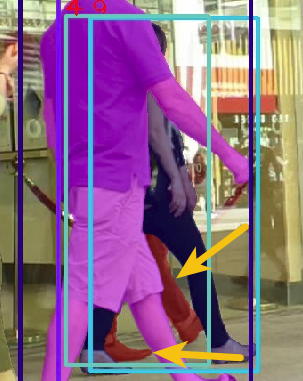}\\
% (a)&(b)&(c)
\end{tabular}
% \vspace*{0.3cm}
\caption{Temporally propagated mask can be helpful in cases of high occlusion. The person with the red mask is tracked only by its limited visible parts (pointed by yellow arrows for the clarity). Input image data from~\cite{mot17_ref}. Best seen in color.}
\vspace*{-0.4cm}
\label{fig:power_of_cutie}
\end{figure}

Tracking multiple objects at once often involves handling challenging occlusions, where only a small part of the subject might be visible, e.g. a leg of a person. Temporally propagated mask can be especially helpful in such cases, when the visible shape can considerably differ from the subject. A visual example is presented in \cref{fig:power_of_cutie}.

We note explicitly that incorporating temporally propagated mask, which also involves temporal coherency, is different than using a static mask coming directly from an image segmentation model (such as SAM~\cite{sam_ref}) independently per each frame. We detail it in \cref{sec:method_mask_management}. To the best of our knowledge, using the mask temporal propagation within the problem of MOT has not been done before. 
% %
% Further, in our work, we do not focus on Single Object Tracking (SOT), Video Object Segmentation (VOS), or Multi-Object Tracking and Segmentation (MOTS). Our work covers specifically the problem of MOT, where we associate newly detected bounding boxes with previously tracked bounding boxes. 
% %
% Hence, we measure the bounding box association performance and not the mask performance, which is already reported in the related works of temporal propagation models [XMem, Cutie ref].
% %
%

Our work covers specifically the problem of MOT, where we associate newly detected bounding boxes with previously tracked bounding boxes. Other challenges, such as Single Object Tracking (SOT), Video Object Segmentation (VOS), or Multi-Object Tracking and Segmentation (MOTS), although relevant, necessitate further modeling and are beyond the scope of this paper. In our work, we measure specifically the performance of bounding box association. For the mask performance, we refer the reader to the related works of temporal propagation models ~\cite{xmem_ref, cutie_ref}.

We evaluate our incorporation of the temporally propagated mask as an association cue against a baseline tracker, showing clear benefits for MOT. 
% in our controlled approach. 
Our tracker is tested on four diverse MOT datasets, achieving the highest performance from tracking-by-detection algorithms on DanceTrack~\cite{dancetrack_ref}, MOT17~\cite{mot17_ref}, SoccerNet-tracking 2022~\cite{soccernet-tracking2022_ref} and KITTI-tracking~\cite{kitti_ref}. These results highlight the advantages of using mask propagation, eliminating the need for per-sequence hyper-parameter tuning. 
% Visual examples of our designed tracker utilizing the temporally propagated mask as an association cue improving over the baseline algorithm are presented in \cref{fig:visual_differences_1,fig:visual_differences_2}.
%
% Visual examples are presented in \cref{fig:visual_differences_1,fig:visual_differences_2}.

Our contribution in this work is summarized as follows:
\begin{enumerate}
    \item We provide an evaluation of the existing mask-based approaches in the MOT domain, demonstrating their unsuitability (\cref{sec:exps_other_mask_methods}).
    \item We propose a novel approach adapting a temporally propagated object segmentation mask as a strong and powerful cue, which, contrarily to a static mask, is incorporated into MOT for the first time. We describe the approach in \cref{sec:method_mask_use}.
    \item We design an MOT tracking algorithm substantiating the idea and incorporating the temporally propagated mask as an association cue between tracklets and detections, detailed in \cref{sec:method_preliminaries,sec:method_mask_management}. The tracker overcomes the limitations of mask-based approaches by performing proper tracklet management and including other important association cues as well as the limitations of the baseline tracking-by-detection approaches, by making the tracking process more robust and generic.
    \item We evaluate and discuss the use of the temporally propagated mask as an association cue with different conditions within a tracking-by-detection algorithm in the ablation study (\cref{sec:exps_abl_stud}). Further, we evaluate our tracker incorporating the mask cue on four different datasets (in \cref{sec:exps_sota_tr_by_det}), comparing it to state-of-the-art tracking-by-detection approaches and demonstrating performance gain thanks to the attentive mask propagation usage, while not tuning per-sequence parameters.
\end{enumerate}

% \begin{figure*}
% \centering
% \begin{tabular}{cccc}

% \includegraphics[height=2.5cm]{images/bt1a.png}&
% \includegraphics[height=2.5cm]{images/bt1b.png}&
% \includegraphics[height=2.5cm]{images/ours1a.png}&
% \includegraphics[height=2.5cm]{images/ours1b.png}
% \\
% % \includegraphics[height=3.0cm]{images/bt2a.png}
% % \includegraphics[height=3.0cm]{images/bt2b.png}&
% % \includegraphics[height=3.0cm]{images/ours2a.png}
% % \includegraphics[height=3.0cm]{images/ours2b.png}
% Frame 319 (baseline)&Frame 401 (baseline)&Frame 319 (McByte)&Frame 401 (McByte)
% \end{tabular}
% \caption{Visual output comparison between the baseline and McByte. With the temporally propagated mask guidance, McByte can handle longer occlusion in the crowd - see the subject with ID 54 on the output of McByte. Input image data from
% ~\cite{mot17_ref}. Best seen in color.}
% \vspace*{-0.3cm}
% \label{fig:visual_differences_1}
% \end{figure*}
\begin{figure*}
\centering
\includegraphics
[width=14cm]
{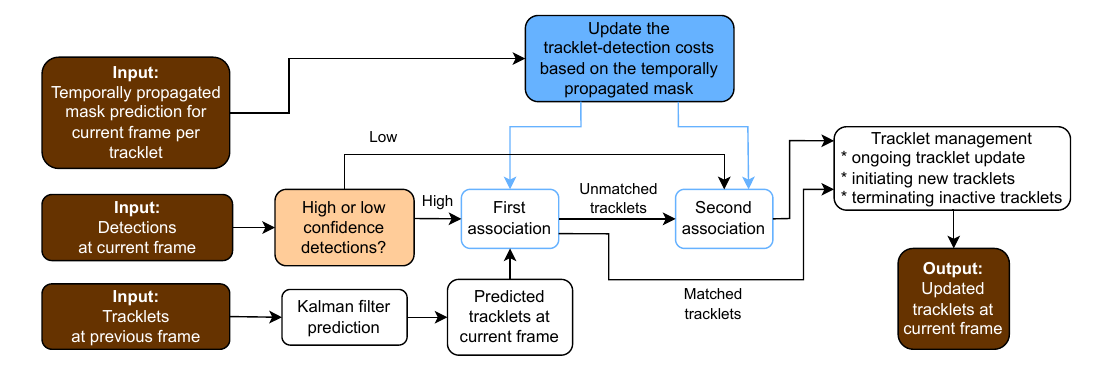}
% \vspace*{0.3cm}
\caption{McByte tracking pipeline with the mask cue guidance. Temporally propagated mask signal is incorporated as an association cue in the tracklet-detection association steps.}
\vspace*{-0.3cm}
\label{fig:diagram_tracking_pipeline}
\end{figure*}

% \vspace*{-0.3cm}

\section{Related Work}
\label{sec:related_work}

Different types of algorithms for tracking multiple objects at the same time have been developed in the community. We provide a brief overview of the existing methods related to our work.

\subsection{Transformer-based and other types of method}
\label{sec:related_work__tranformers}

Transformer-based methods~\cite{motip_ref,motrv2_ref,memotr_ref, motr_ref} use attention mechanisms to learn tracking trajectories and object associations from training data, following an end-to-end approach. These methods perform well on video sequences where subjects remain mostly on the scene, such as DanceTrack~\cite{dancetrack_ref}, but struggle with sequences where subjects frequently enter and exit, like in the MOT17 dataset~\cite{mot17_ref}, despite being designed for MOT. For this reason, we do not directly compare our method with transformer-based approaches in the main paper, though their results are listed together with the other approaches and ours in \cref{sec:sota_transformers_sushi}.

Other types of MOT methods have also been proposed in the community, such as global optimization (offline) methods~\cite{sushi_ref} or joint detection and tracking methods~\cite{fairmot_ref,relationtrack_ref}. Analogously to the transformer-based methods, they are not directly comparable and we list them for reference in \cref{sec:sota_transformers_sushi}.

\subsection{Tracking-by-detection methods}
\label{sec:related_work__tr_by_det}

The tracking-by-detection approach detects objects in each video frame and associates them into tracklets, linking the same objects across frames.

ByteTrack~\cite{bt_ref} is a powerful tracking-by-detection algorithm that uses the YOLOX~\cite{yolox_ref} detector, associating tracklets based on the intersection over union (IoU) between bounding boxes of tracklets and new detections. It provides strong tracklet management and serves as a solid baseline for MOT. Several works have built on ByteTrack. OC-SORT~\cite{ocsort_ref} enhances state estimation by computing virtual trajectories during occlusion. StrongSORT~\cite{strongsort_ref} adds re-identification (re-ID) features as cues, camera motion compensation, and NSA Kalman filter~\cite{nsa_kf_ref}. C-BIoU~\cite{cbiou_ref} extends the association process by buffering (enlarging) bounding boxes, while HybridSORT~\cite{hybridsort_ref} adds cues like confidence modeling and height-modulated IoU alongside existing strong cues.

Although these algorithms perform well on popular MOT datasets, they are highly sensitive to parameters. ByteTrack, for instance, explicitly tunes parameters like high-confidence detection 
% and tracklet initialization 
thresholds per test sequence, as noted on its GitHub page\footnote{\url{https://github.com/ifzhang/ByteTrack}, see: Test on MOT17.} and in the code\footnote{\url{https://github.com/ifzhang/ByteTrack/blob/main/yolox/evaluators/mot_evaluator.py}, lines 146-157 (at the moment of submitting this paper)}, which affects the tracking performance. Extensions of ByteTrack~\cite{strongsort_ref,ocsort_ref,deepocsort_ref, hybridsort_ref}, sharing the same code also rely on per-sequence parameter tuning. This tuning process is costly and impractical for larger datasets like DanceTrack~\cite{dancetrack_ref}, SoccerNet-tracking~\cite{soccernet-tracking2022_ref}, and KITTI-tracking~\cite{kitti_ref}, as well as for generalizing across datasets. In contrast, our method, incorporating the temporally propagated segmentation mask as an association cue, avoids per-sequence tuning, making it more robust and generic, as we further demonstrate in \cref{sec:exps_sota_tr_by_det}.

% \begin{figure*}
% \centering
% % \includegraphics[height=3.0cm]{images/diagram1_mask_management.drawio.png}
% % \includegraphics[width=12cm]{images/Copy of Copy of diagram1_mask_management.drawio(1).png}
% \includegraphics
% [width=14cm]
% {images/CVPR25__diagram2_tracking_pipline.drawio.pdf}
% % \vspace*{0.3cm}
% \caption{McByte tracking pipeline with the mask cue guidance. Temporally propagated mask signal is incorporated as an association cue in the tracklet-detection association steps.}
% \vspace*{-0.3cm}
% \label{fig:diagram_tracking_pipeline}
% \end{figure*}
\begin{figure*}
\centering
\begin{tabular}{cccc}

\includegraphics[height=2.5cm]{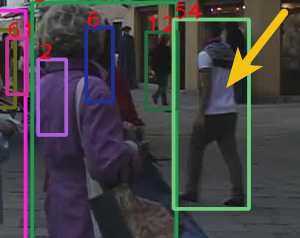}&
\includegraphics[height=2.5cm]{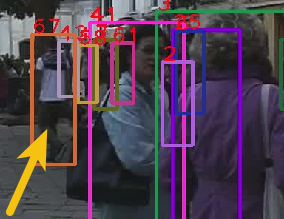}&
\includegraphics[height=2.5cm]{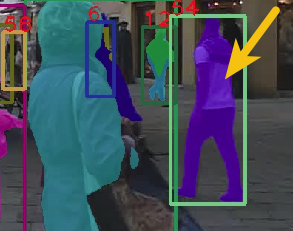}&
\includegraphics[height=2.5cm]{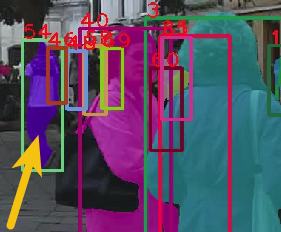}
\\
Frame 319 (baseline)&Frame 401 (baseline)&Frame 319 (McByte)&Frame 401 (McByte)
\end{tabular}
\caption{Visual output comparison between the baseline and McByte. With the temporally propagated mask guidance, McByte can handle longer occlusion in the crowd - see the subject with ID 54 on the output of McByte. Input image data from
~\cite{mot17_ref}. Best seen in color.}
\vspace*{-0.3cm}
\label{fig:visual_differences_1}
\end{figure*}

\subsection{Segmentation mask models}
\label{sec:related_work__seg_mask_models}

\textbf{Mask creation.} The Segment-Anything Model (SAM)~\cite{sam_ref} is a highly effective image segmentation model trained on a massive dataset, delivering impressive segmentation outputs. Recently, SAM 2~\cite{sam2_ref} was introduced, enhancing SAM’s performance and enabling video-level segmentation across entire sequences. However, once tracking starts, SAM 2 cannot track new objects, limiting its use in complete MOT tasks. Later in this section, we discuss a SAM~2-based bounding box tracker combined with Grounding Dino~\cite{gr_dino_ref} (see: \textit{Mask-based tracking systems}).

\textbf{Mask temporal propagation.} XMem~\cite{xmem_ref} is a mask temporal propagation model for video object segmentation (VOS), based on the Atkinson-Shiffrin Memory Model~\cite{atk_shf_mem_model_ref}, enabling long-term tracking of segmentation masks. Its successor, Cutie~\cite{cutie_ref}, improves segmentation by incorporating object encoding from mask memory and better distinguishing the object from the background. Note that image segmentation models like SAM~\cite{sam_ref} only constitute the creation of the initial mask at the designated frames. Mask temporal propagation models then aim to infer the referred mask over the next frames in a video sequence.

% While useful for re-identifying subjects across the adjacent frames, both XMem and Cutie are not suitable for MOT as they do not involve bounding boxes and can provide inaccurate mask predictions as listed in their performance on video object segmentation~\cite{xmem_ref, cutie_ref}. 
While generally powerful in their target domains, both XMem and Cutie are not directly suitable for MOT as they do not involve bounding boxes and might provide inaccurate mask predictions as listed in their performance on video object segmentation~\cite{xmem_ref, cutie_ref}. 
Therefore, we propose a novel MOT algorithm that combines temporally propagated masks with bounding boxes, improving tracking performance. Our ablation studies in \cref{sec:exps_abl_stud} show that using the mask in a controlled manner, 
% alongside 
and along with
other MOT mechanisms, is more effective than relying solely on the mask signal.

\textbf{Mask-based tracking systems.} Segmentation mask models have already been used to build tracking systems. DEVA~\cite{deva_ref}, which enhances XMem~\cite{xmem_ref}, proposes decoupled video segmentation and bi-directional propagation, and builds a tracking system with boxes and masks. However, MOT requires a robust tracklet management and an occlusion handling system, which is not available within DEVA. 

Grounded SAM 2\footnote {Method not officially published, yet with implementation code available: \url{https://github.com/IDEA-Research/Grounded-SAM-2}} is a tracking system combining Grounding Dino~\cite{gr_dino_ref} and SAM 2~\cite{sam2_ref} for video object segmentation, designed to track bounding boxes and maintain object IDs. However, SAM 2’s inability to add new objects once tracking begins makes it unsuitable for MOT. It tracks objects in segments, but merging those segments is not always successful.

MASA~\cite{masa_ref} is a mask feature-based adapter~\cite{adapters_ref,adapters_cv_ref} trained with SAM~\cite{sam_ref}, offering inference modes for video segmentation and object tracking. In tracking, MASA provides features for matching object detections across frames. However, it struggles with longer occlusions and missing or incorrect detections. While MASA has been tested on a few MOT datasets~\cite{masa_ref}, the mentioned limitations weaken its performance and reduce its generalizability across varied datasets.

Existing mask-based tracking algorithms, though trained on large datasets for extracting semantic features, fail to handle challenges like occlusion and track initiation. We discuss it more in detail in \cref{sec:exps_other_mask_methods}, where we show that their performance on MOT tasks is significantly lower than other methods. Our proposed MOT tracker incorporates the temporally propagated mask as an association cue while leveraging powerful mechanisms like tracklet management and additional cues (e.g., bounding box position, motion). As we demonstrate in \cref{sec:exps_other_mask_methods}, our approach performs considerably better than existing mask-based systems, even when they are adapted to MOT format and evaluation.

\section{Proposed method}
\label{sec:method}

\subsection{Preliminaries}
\label{sec:method_preliminaries}

% \textbf{\textcolor{red}{[SAY MORE LIKE BASELINE. SURE, YOU CAN MENTION BYTETRACK ONCE, BUT DO NOT VALORIZE IT TOO MUCH. AND REFER TO IT AS BASELINE] }}

% Multi-object tracking (MOT) involves tracking objects of interest detected in a video sequence. So called tracklets are formed to link the detections over the consecutive frames. In case of tracking-by-detection, a common practice~\cite{bt_ref,ocsort_ref, deepocsort_ref,cbiou_ref,strongsort_ref,hybridsort_ref} is associating already existing tracklets built over the preceding frames with new detections of the currently processed frame. Based on applied association cues, such as object position information and motion information, a cost matrix is built including potential cost (also referred to as distance) between each considered tracklet-detection association pair. The association problem is then considered as a bipartite matching problem and Hungarian matching algorithm~\cite{hungarianalg_ref} is deployed to match the tracklets with detections towards extending the tracklets, while optimizing for the lowest total cost of all pairs. The pairs with a cost values above the defined match threshold are excluded from the association.
%
Multi-object tracking (MOT) involves tracking detected objects across video frames by forming tracklets that link detections over consecutive frames. In tracking-by-detection methods~\cite{bt_ref,ocsort_ref, deepocsort_ref,cbiou_ref,strongsort_ref,hybridsort_ref}, existing tracklets from previous frames are associated with new detections in the current frame. This process uses association cues such as object position and motion to build a cost matrix representing potential tracklet-detection matches. The association problem considered as the bipartite matching problem is then solved using the Hungarian algorithm~\cite{hungarianalg_ref}, which matches tracklets with detections to extend them while minimizing total cost. Pairs with costs above the pre-defined matching threshold are excluded.

% Following the practices from our baseline, ByteTrack~\cite{bt_ref}, 
% % In case of ByteTrack~\cite{bt_ref}, which we use as our baseline, 
% new detections are split into two groups with higher and lower detection confidence scores and handled in separate association steps. For its association process, the baseline primarily uses intersection over union (IoU). Tracklet position bounding boxes at the current frame are predicted using Kalman Filter~\cite{kf_ref} and are compared to the detection bounding boxes based on the IoU score. Note that IoU $\in [0, 1]$. Cost matrix entries are then filled with 1-IoU score for each tracklet-detection pair. Besides the association part, tracklet management is carefully performed including initiating new tracklets, updating ongoing tracklets and terminating inactive tracklets. We refer the reader to the baseline paper~\cite{bt_ref} for more details if needed.

In our baseline, ByteTrack~\cite{bt_ref}, new detections are split into high and low confidence groups, handled separately in the association process. The baseline uses intersection over union (IoU) as the primary association metric. Tracklet positions are predicted using a Kalman Filter~\cite{kf_ref} and compared to detection bounding boxes using IoU scores. Cost matrix entries are filled with 1-IoU for each tracklet-detection pair. Further, tracklet management includes initiating, updating, and terminating tracklets. For more details, we refer the reader to the ByteTrack paper~\cite{bt_ref}.

In our work, we study a temporally propagated segmentation mask as a powerful association cue for MOT. We extend the baseline algorithm~\cite{bt_ref} and combine the mask information with bounding box information to create our novel \textbf{m}asked-\textbf{c}ued algorithm, which we call McByte. \cref{fig:diagram_tracking_pipeline} shows the overview of our tracking pipeline pointing to where the temporally propagated mask is involved as an association cue. 
% We note that we consider the problem of MOT, where we associate the bounding boxes of tracklets and detections, and not the problems of Single Object Tracking (SOT), Video Object Segmentation (VOS) or Multi-Object Tracking and Segmentation (MOTS). We measure the performance of the bounding box association and not the performance of the mask, which is already reported in the related works [XMem and Cutie refs]. 
We note that we consider the problem of MOT, where we perform and evaluate the association between the bounding boxes of detections and tracklets.

In the next sections, we describe the creation and handling of the mask within our MOT tracking algorithm (\cref{sec:method_mask_management}) and our regulated use of the temporally propagated mask as an association cue (\cref{sec:method_mask_use}).

\subsection{Mask creation and handling}
\label{sec:method_mask_management}

Since the temporally propagated mask as a cue has to facilitate associating a tracklet with the correct detection, it must be well correlated with the tracklets and handled accordingly. This process is originally not straightforward and thus we design the following approach of mask handling to synchronise it with the processed tracklets.

During tracking in McByte, each tracklet gets its own mask, which is then temporally propagated across frames to update the mask predictions. At first, we use an image segmentation model to create a segmentation mask for each new tracklet. It is performed only for a newly appeared object and to initialize a new mask. 

Separately, during the next frames, a temporal propagator is used to infer the updated mask positions, while aiming to keep the spatio-temporal consistency of the mask. The propagator predictions are analyzed and used in the tracklet-detection association process, as detailed in \cref{sec:method_mask_use}. We manage the temporally propagated masks in sync with the tracklet management system, creating new masks for new tracklets and removing them when a tracklet is terminated. 

We provide more detailed information and a diagram including both mask creation and mask temporal propagation in \cref{sec:components_in_detail}.
% Visual examples of the masks being used within our tracker are shown in \cref{fig:visual_differences_1,fig:visual_differences_2}.

% \begin{figure}
% \centering
% % \includegraphics[height=3.0cm]{images/diagram1_mask_management.drawio.png}
% % \includegraphics[width=12cm]{images/Copy of Copy of diagram1_mask_management.drawio(1).png}
% \includegraphics
% [width=8cm]
% {images/mm1_and_mm2.drawio-7.pdf}
% % \vspace*{0.3cm}
% \caption{Cases showing the differences in $mm_{1}$ and $mm_{2}$ (\cref{sec:method_mask_use}) values of a mask (in blue) within a bounding box. The most optimal case for the mask to provide a good guidance is the second one from the left, where
% % when the mask is sufficiently big and when it is covered by a bounding box as much as possible, i.e 
% both $mm_{1}$ and $mm_{2}$ are as close to $1$ as possible.}
% % \vspace*{-0.5cm}
% \label{fig:mm1_mm2}
% \end{figure}

\subsection{Regulated use of the mask}
\label{sec:method_mask_use}

% A temporally propagated segmentation mask of a tracked object can be a powerful association cue as long as it is used properly. The mask prediction coming from a temporal mask propagator can sometimes be incorrect as demonstrated on the video segmentation results of the related works~\cite{xmem_ref,cutie_ref}, and thus unreliable. This implies a need for the regulated use of the mask as a cue.
%
A temporally propagated (TP) segmentation mask can be a strong association cue if used correctly. However, mask predictions from the temporal propagator might sometimes be incorrect, as seen in related works~\cite{xmem_ref,cutie_ref} and thus unreliable, making it essential to regulate the mask's use as a cue.

% During the association process, we use the mask to update the entries of the cost matrix, where the cost values between tracklets and detections are ambiguous. By ambiguity, we mean cases where a tracklet could be potentially matched to more than one detection or where a detection could be matched to more than one tracklet. Such an ambiguity might frequently stem from the matches being based on the IoU score. When the tracked objects are close to each other, their bounding boxes significantly overlap resulting is similar IoU scores and cost values. Therefore, if the IoU is below the matching threshold for more than one tracklet-detection bounding box pair for the same tracklet or detection, we consider it as an ambiguity.
%
In the association process, we update the cost matrix entries using the TP mask, particularly in cases of ambiguity, where a tracklet could match multiple detections or vice versa. Ambiguity often arises from IoU-based matches when tracked objects are close, causing significant overlap in bounding boxes and similar IoU scores. If IoU is below the matching threshold for more than one tracklet-detection pair,
% of a tracklet or of a detection
we treat it as ambiguity.

\begin{figure}
\centering
\includegraphics
[width=7cm]
{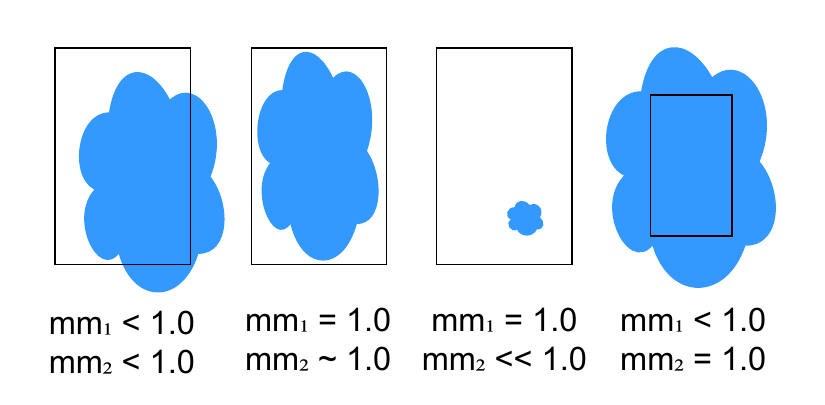}
% \vspace*{0.3cm}
\caption{Cases showing the differences in $mm_{1}$ and $mm_{2}$ (\cref{sec:method_mask_use}) values of a temporally propagated mask (in blue) within a bounding box. The most optimal case for the mask to provide a good guidance is the second one from the left, where
% when the mask is sufficiently big and when it is covered by a bounding box as much as possible, i.e 
both $mm_{1}$ and $mm_{2}$ are as close to $1$ as possible.}
\vspace*{-0.3cm}
\label{fig:mm1_mm2}
\end{figure}

For each potential ambiguous tracklet-detection match, we apply a strategy consisting of the following conditions.
\begin{enumerate}
\item We check if the considered tracklet's TP mask is actually visible on the scene. Subjects can be entirely occluded resulting in no mask prediction at the current frame.
\item We check if the TP mask prediction is confident enough, i.e. if the average mask probability for the given object from the mask propagator is above the set mask confidence threshold. 
\end{enumerate}
Further, we compute two key ratios between the TP mask and detection bounding box:
\begin{itemize}
\item the bounding box coverage of the mask, referred to as mask match no. 1, $mm_{1}$:
\end{itemize}
\begin{equation}
     mm^{i,j}_{1} = \frac{|pix(mask(tracklet_{i})) \cap pix(bbox_{j})|}{|pix(mask(tracklet_{i}))|} 
  \end{equation}
\begin{itemize}
\item the mask fill ratio of the bounding box, referred to as mask match no. 2, $mm_{2}$:
\end{itemize}
\begin{equation}
    mm^{i,j}_{2} = \frac{|pix(mask(tracklet_{i})) \cap pix(bbox_{j})|}{|pix(bbox_{j})|} 
  \end{equation}

where $pix(\cdot)$ denotes pixels of the mask or within the bounding box, and $mask(\cdot)$ denotes the TP mask assigned to the tracklet. $|\cdot|$ denotes the cardinality of the set. Note that all $mm_{1}, mm_{2} \in [0, 1]$. In \cref{fig:mm1_mm2}, we show how $mm_{1}$ and $mm_{2}$ can vary depending on the TP mask and bounding box position. We discuss it more in detail in \cref{sec:components_in_detail}. Going further with the conditions:
\begin{enumerate}
\setcounter{enumi}{2}
\item We check if the mask fill ratio of the bounding box $mm_{2}$ occupies a significant portion of the bounding box.
\item We check if the bounding box coverage of the mask $mm_{1}$ is sufficiently high. 
\end{enumerate}
Only if all these conditions hold, we update the tracklet-detection association in the cost matrix using the following formula:
\begin{equation}
     costs^{i,j} = costs^{i,j} - mm^{i,j}_{2}
     \label{eq:cost_matrix_update}
\end{equation}
where $costs^{i,j}$ denotes the cost between tracklet \textit{i} and detection \textit{j}. With this fusion of the available information, we consider both modalities, TP masks and bounding boxes to enhance the association process. The updated cost matrix, enriched by the mask signal, is passed to the Hungarian matching algorithm to find optimal tracklet-detection pairs.

% By following the conditions \textit{1}-\textit{4.}, we ensure that the mask cue is actually controlled and that the cost matrix is updated only if the mask seems reliable. We show the impact of each condition (as well as the lack of any condition) in the ablation study in \cref{sec:exps_abl_stud}, \cref{tab:ablation_mask_constraints_gradual}. More detailed explanations of these conditions are included in the supplementary material, Appendix F. 
%
Following conditions \textit{1}-\textit{4.} ensures that the TP mask cue is controlled, and the cost matrix is updated only when the mask is reliable. The impact of each condition, along with the absence of any, is demonstrated in the ablation study (\cref{sec:exps_abl_stud}, \cref{tab:ablation_mask_constraints_gradual}). Further details on these conditions are included in \cref{sec:components_in_detail}.

% Further, the direct use of $mm_{1}$ to influence the cost matrix could be misleading, because more than one mask can be completely included within one, the same bounding box, resulting in $mm_{1}=1.0$ for all these masks. Hence, we use $mm_{1}$ only as a gating condition and use $mm_{2}$ to influence the cost matrix.  
% 
Further, using $mm_{1}$ directly to influence the cost matrix could be misleading, as multiple TP masks could fully fit within the same bounding box, all resulting in $mm_{1}=1.0$. Therefore, we use $mm_{1}$ only as a gating condition and  $mm_{2}$ to influence the cost matrix.

% Since our baseline is designed to work exactly on the bounding boxes, which is suitable for MOT, we do not change the idea of running the Hungarian matching algorithm over the cost matrix. Instead, we carefully incorporate the mask signal to influence the cost and to improve the association process. In our ablation study in \cref{sec:exps_abl_stud}, we demonstrate that using bounding boxes with mask is actually more beneficial than using the mask on its own. We also show the benefits of adding the mask signal as an association cue over the pure bounding box-based association. 
%
Our baseline is optimized for bounding boxes, so we retain the use of the Hungarian matching algorithm over the cost matrix, but we carefully incorporate the TP mask signal to enhance the association process. In the ablation study (\cref{sec:exps_abl_stud}), we show that combining bounding boxes with the mask is more effective than using the mask alone, and demonstrate the advantages of incorporating the TP mask signal as an additional association cue over a purely bounding box-based approach.

In \cref{fig:visual_differences_1}, we show that McByte can handle challenging scenarios such as long-term occlusion in crowd 
% and association of ambiguous boxes 
due to the TP mask signal in the controlled manner as an association cue. More visual examples are available in \cref{sec:components_in_detail}.

\subsection{Handling camera motion}
\label{sec:method_mcbyte_with_cmc}

% When the video sequence camera is moving, the bounding boxes of tracklets and detections might be less accurate due to the change of object motion and blurry objects. As in our tracklet-detection association process we fuse both the mask information and bounding box information, we add camera motion compensation (CMC) to improve the quality of estimated bounding boxes. We use similar approach as in the literature~\cite{strongsort_ref,deepocsort_ref}. We provide more details about the applied CMC in the supplementary material, Appendix F. The final version of McByte includes CMC and we show its impact in the ablation study in \cref{sec:exps_abl_stud},  \cref{tab:ablation_mask_constraints_gradual}.
%
When the camera moves, tracklet and detection bounding boxes may become less accurate due to object motion and blurring. To address this, we integrate camera motion compensation (CMC) into our process (which fuses temporally propagated mask and bounding box information) to enhance the accuracy of bounding box estimates. This approach follows existing methods~\cite{strongsort_ref,deepocsort_ref}. More details on CMC are provided in \cref{sec:components_in_detail}. The final McByte version includes CMC, and its impact is shown in the ablation study in \cref{sec:exps_abl_stud},  \cref{tab:ablation_mask_constraints_gradual}.

\section{Experiments and discussion}
\label{sec:experiments_discussion}

\subsection{Implementation details}
\label{sec:exps_impl_details}

For object detections, we follow the practices of our baseline~\cite{bt_ref} and use the YOLOX~\cite{yolox_ref} detector pretrained on the relevant dataset, unless stated otherwise. As in baseline, detections are divided into high and low confidence sets based on a defined threshold. However, on the contrary to the baseline which considers different values per video sequence, we always use the same threshold for all sequences. We fix it as 0.6, which is the default value of our baseline, i.e. when the baseline is evaluated on datasets without tuning per sequence. 
% For KITTI-tracking (see \cref{sec:exps_datasets_metrics}), we adjust the threshold to 0.5 due to different detection quality. 
%
% In both our baseline and McByte, the new tracklet initialization threshold is set as 0.1 higher than the detection confidence threshold.
% 
% All other parameters match those in the baseline~\cite{bt_ref}.

For mask creation, we use SAM~\cite{sam_ref}, ensuring fair comparison with related works, with the vit\_b model and weights from SAM's authors. As a mask temporal propagator, we use Cutie~\cite{cutie_ref} with the weights provided by the authors, Cutie base mega. Note that we use the image segmentation model (SAM) only to create and initialize the masks for the newly appeared objects at the scene. For the association cue between the tracklets and detections we process the propagated mask from the mask temporal propagator (Cutie). All experiments are conducted on an Nvidia A100 GPU with 40 GB of memory.

\vspace*{-0.08cm}

\subsection{Datasets and evaluation metrics}
\label{sec:exps_datasets_metrics}

We 
% focus on tracking people and 
evaluate McByte on four person tracking datasets with diverse characteristics, to demonstrate the generalizability of our method. We present results on DanceTrack~\cite{dancetrack_ref}, MOT17~\cite{mot17_ref}, SoccerNet-tracking 2022~\cite{soccernet-tracking2022_ref}, and KITTI-tracking~\cite{kitti_ref}, while using detection sources per dataset as per community practices for fair comparison.

DanceTrack~\cite{dancetrack_ref} features people performing dances with highly non-linear motion and subtle camera movements, while the number of individuals stays mostly constant. We use YOLOX pre-trained on DanceTrack for detections, following our baseline~\cite{bt_ref}.

MOT17~\cite{mot17_ref} involves tracking people in public spaces under varying conditions, such as lighting, pedestrian density, and camera stability. We use YOLOX pre-trained on MOT17 for detections as provided by the the baseline~\cite{bt_ref}.

SoccerNet-tracking 2022~\cite{soccernet-tracking2022_ref} contains soccer match videos, where players move rapidly and look alike in the same team. Camera movement is always present and oracle detections are provided by the dataset authors.

KITTI-tracking~\cite{kitti_ref} captures 
% pedestrian and vehicle 
scenes from car view, with varied pedestrian density and frequent camera movement. It also considers another class, car, which can help to assess the generalizability of the tracking method. Following community practices~\cite{ocsort_ref,strongsort_ref}, we use PermaTr~\cite{permatr_ref} detections.

We report three standard MOT metrics: HOTA~\cite{hota_ref}, IDF1~\cite{idf1_ref} and MOTA~\cite{mota_ref}, focusing on HOTA and IDF1 for evaluating the tracking performance. IDF1 measures the tracking quality and identity preservation, while HOTA includes association, detection and localization. MOTA primarily measures detection quality and we report it for the result completeness. Higher values in these metrics indicate better performance.
% Limited number of submissions is allowed on the test set evaluation servers of MOT17~\cite{mot17_ref} and KITTI-tracking~\cite{kitti_ref}.
% As we focus on the problem of MOT in this work, we report the bounding box association performance using the aforementioned metrics.

\begin{table}
  \centering
  {\small{
  \begin{tabular}{@{}lccc@{}}
    \toprule
    Method & HOTA & IDF1 & MOTA \\
    \midrule
    basline~\cite{bt_ref}: no mask & 47.1 & 51.9 & 88.2 \\
    a1: either mask or no assoc. & 48.6 & 44.4 & 80.8 \\ 
    a2: either mask or IoU for assoc. & 45.3 & 41.5 & 82.2 \\ 
    % ablation 2b \textcolor{magenta}{2a XOR 2b to be removed} & 44.3 & 40.0 & 84.6 \\ 
    a3: IoU and mask if ambiguity & 56.6 & 57.0 & 89.5 \\ 
    a4: a3 + mask confidence & 57.3 & 57.7 & 89.6 \\ 
    a5: a4 + $mm_{2}$ & 58.8 & 60.1 & 89.6 \\ 
    a6: a5 + $mm_{1}$ & 62.1 & 63.4 & 89.7 \\
    % ablation 3e & 62.1 & 63.4 & 89.7 \\ 
    McByte: a6 + cmc & \textbf{62.3} & \textbf{64.0} & \textbf{89.8} \\
    \bottomrule
  \end{tabular}
  }}
  \caption{Ablation study on DanceTrack~\cite{dancetrack_ref} validation set listing the effects of the imposed constraints on using the temporally propagated mask as an association cue.}
  \label{tab:ablation_mask_constraints_gradual}
  % \vspace*{-0.3cm}
\end{table}
\begin{table}
  \centering
  {\small{
  \begin{tabular}{@{}lccc@{}}
    \toprule
    Method & HOTA & IDF1 & MOTA \\
    \midrule
    % \multicolumn{4}{c}{DanceTrack val} \\
    % \midrule
    Baseline, DanceTrack val                & 47.1 & 51.9 & 88.2 \\
    % baseline + M (=abl.3d)  & 62.1 & 63.4 & 89.7 \\ 
    % baseline + LC           & 54.4 & 55.9 & 59.7 \\ 
    McByte, DanceTrack val                  & \textbf{62.3} & \textbf{64.0} & \textbf{89.8} \\ 
    % \textcolor{gray}{baseline + C}            & \textcolor{gray}{53.7} & \textcolor{gray}{54.3} & \textcolor{gray}{89.6} \\ 
    % \textcolor{gray}{baseline + C + M}        & \textcolor{gray}{61.0} & \textcolor{gray}{61.9} & \textcolor{gray}{89.8} \\
    \midrule
    % \multicolumn{4}{c}{SoccerNet-tracking 2022 test} \\
    % \midrule
    % baseline                & \textcolor{lightgray}{71.5} & \textcolor{lightgray}{-} & \textcolor{lightgray}{94.6} \\
    % baseline our run        & 72.1 & 75.3 & 94.5 \\
    Baseline, SoccerNet-tracking 2022 test  & 72.1 & 75.3 & 94.5 \\
    % baseline + M (=abl.3d)  & 82.9 & 77.1 & 96.6 \\
    % baseline + C            & 84.8 & 79.7 & 96.9 \\ 
    McByte, SoccerNet-tracking 2022 test    & \textbf{85.0} & \textbf{79.9} & \textbf{96.8} \\
    \midrule
    % \multicolumn{4}{c}{MOT17 val} \\
    % \midrule
    Baseline, MOT17 val                     & 68.4 & 80.2 & 78.2 \\
    % baseline + M (=abl.3d)  & 69.2 & 81.3 & 78.3 \\
    % baseline + C            & 69.7 & 82.2 & 79.0 \\ 
    McByte, MOT17 val                       & \textbf{69.9} & \textbf{82.8} & \textbf{78.5} \\
    \midrule
    % \multicolumn{4}{c}{KITTI-tracking test} \\
    % \midrule
    Baseline, KITTI-tracking test           & 54.3 & - & 63.7 \\
    McByte, KITTI-tracking test             & \textbf{57.0} & - & \textbf{68.9} \\
    \bottomrule
  \end{tabular}
  }}
  \caption{Ablation study comparing McByte with the baseline~\cite{bt_ref} on four different datasets. As SoccerNet-tracking 2022~\cite{soccernet-tracking2022_ref} and KITTI-tracking~\cite{kitti_ref} (pedestrian) do not contain validation set split, we report the results on the test sets. KITTI evaluation server does not provide IDF1 scores.}
  \label{tab:ablation_mask_cmc}
  \vspace*{-0.3cm}
\end{table}

\subsection{Ablation studies}
\label{sec:exps_abl_stud}

We perform an ablation study to demonstrate the impact of incorporating the temporally propagated (TP) mask as an association cue along with the conditions discussed in \cref{sec:method_mask_use}. We evaluate the following variants:
\begin{itemize}
    \item a1: Uses only the TP mask signal for association if the mask is visible for the given tracklet, without ambiguity checks (\cref{sec:method_mask_use}). The value of $mm^{i,j}_{2}$ is directly assigned to $costs^{i,j}$ in \cref{eq:cost_matrix_update}. No association occurs if there is no mask.
    \item a2: Similar to a1, but if the TP mask is unavailable, intersection over union (IoU) scores are used for association, as in the baseline~\cite{bt_ref}.
    \item a3: Adds an ambiguity check (\cref{sec:method_mask_use}). If the TP mask is visible, mask and bounding box information are fused as shown in \cref{eq:cost_matrix_update}. If no mask is available, IoU scores are used as in the baseline.
    \item a4: Builds on a3, incorporating the mask confidence check (condition 2 in \cref{sec:method_mask_use}).
    \item a5: Extends a4 by adding the $mm_{2}$ value check from condition 3 in \cref{sec:method_mask_use}.
    \item a6: Further extends a5 with the $mm_{1}$ value check from condition 4 in \cref{sec:method_mask_use}.
\end{itemize}

\begin{table}
  \centering
  {\small{
  \begin{tabular}{@{}lccc@{}}
    \toprule
    Method & HOTA & IDF1 & MOTA \\
    % \midrule
    % \multicolumn{4}{c}{\textcolor{magenta}{Mind extended version! With}} \\
    % \multicolumn{4}{c}{\textcolor{magenta}{Transformers and SUSHI. BMVC Suppl. Mat}} \\
    \midrule
    ByteTrack~\cite{bt_ref}             & 47.7 & 53.9 & 89.6 \\
    OC-SORT~\cite{ocsort_ref}           & 55.1 & 54.9 & 92.2 \\
    Deep OC-SORT~\cite{deepocsort_ref}  & 61.3 & 61.5 & 92.3 \\
    % C-BIoU~\cite{cbiou_ref}             & \textcolor{lightgray}{60.6} & \textcolor{lightgray}{61.6} & \textcolor{lightgray}{91.6} \\
    C-BIoU~\cite{cbiou_ref} *           & 45.8 & 52.0 & 88.4 \\
    StrongSORT++~\cite{strongsort_ref}  & 55.6 & 55.2 & 91.1 \\
    Hybrid-SORT~\cite{hybridsort_ref}   & 65.7 & 67.4 & 91.8 \\
    % Ours (McByte)            & 66.5 & \textbf{68.2} & 92.7 \\
    McByte (ours)                       & \textbf{67.1} & \textbf{68.1} & \textbf{92.9} \\
    \bottomrule
  \end{tabular}
  }}
  \caption{Comparing McByte with state-of-the-art tracking-by-detection algorithms on DanceTrack test set~\cite{dancetrack_ref}. 
  % *We reproduce C-BIoU and report the best result obtained.
  }
  \label{tab:sota_dancetrack_test}
\end{table}

\begin{table}
  \centering
  {\small{
  \begin{tabular}{@{}lccc@{}}
    \toprule
    Method & HOTA & IDF1 & MOTA \\
    % \midrule
    % \multicolumn{4}{c}{\textcolor{magenta}{Mind extended version! With}} \\
    % \multicolumn{4}{c}{\textcolor{magenta}{Transformers and SUSHI. BMVC Suppl. Mat}} \\
    \midrule
    \multicolumn{4}{c}{With parameter tuning per sequence} \\
    \midrule
    \textcolor{gray}{ByteTrack~\cite{bt_ref}}             & \textcolor{gray}{63.1} & \textcolor{gray}{77.3} & \textcolor{gray}{80.3} \\
    % BoT-SORT~\cite{botsort_ref}           & 65.0 & 80.2 & 80.5 \\
    \textcolor{gray}{StrongSORT++~\cite{strongsort_ref}}  & \textcolor{gray}{64.4} & \textcolor{gray}{79.5} & \textcolor{gray}{79.6} \\
    % \textcolor{violet}{C-BIoU~\cite{cbiou_ref}}          & \textcolor{violet}{64.1} & \textcolor{violet}{79.7} & \textcolor{violet}{81.1} \\
    % C-BIoU~\cite{cbiou_ref} *           & TO & BE & RUN \\
    % ImprAsso~\cite{imprasso_ref}        & \textbf{66.4} & \textbf{82.1} & \textbf{82.2} \\
    \textcolor{gray}{OC-SORT~\cite{ocsort_ref}}           & \textcolor{gray}{63.2} & \textcolor{gray}{77.5} & \textcolor{gray}{78.0} \\
    \textcolor{gray}{Deep OC-SORT~\cite{deepocsort_ref}}  & \textcolor{gray}{64.9} & \textcolor{gray}{80.6} & \textcolor{gray}{79.4} \\
    \textcolor{gray}{Hybrid-SORT~\cite{hybridsort_ref}}   & \textcolor{gray}{64.0} & \textcolor{gray}{78.7} & \textcolor{gray}{79.9} \\
    % \midrule
    % \multicolumn{4}{c}{Without parameter tuning per sequence} \\
    % \midrule
    % \multicolumn{4}{c}{\textbf{\textcolor{magenta}{ByteTrack: with our without interpolation?}}} \\
    % \multicolumn{4}{c}{\textcolor{magenta}{Ours: WITH INTERPOLATION}} \\
    % \multicolumn{4}{c}{\textcolor{magenta}{ByteTrack with no param tuning reported}} \\
    % \multicolumn{4}{c}{\textcolor{magenta}{by GHOST and SUSHI: WITHOUT INTERPOLATION}} \\
    \midrule
    % \multicolumn{4}{c}{No implementation provided} \\
    \multicolumn{4}{c}{Without parameter tuning per sequence} \\
    \midrule
    ByteTrack~\cite{sushi_ref} & 62.8 & 77.1 & 78.9 \\
    C-BIoU~\cite{cbiou_ref} *          & 62.4 & 77.1 & 79.5 \\
    % \midrule
    % \multicolumn{4}{c}{Without parameter tuning per sequence} \\
    % \midrule
    % \textcolor{red}{[REMOVE IT]} \textcolor{magenta}{ByteTrack (with interp.)} & 63.2 & 77.5 & 79.7 \\
    % Ours (McByte)                    & 63.1 & 78.0 & 79.7 \\
    McByte (ours)                      & \textbf{64.2} & \textbf{79.4} & \textbf{80.2} \\
    \bottomrule
  \end{tabular}
  }}
  \caption{Comparing McByte with state-of-the-art tracking-by-detection algorithms on MOT17 test set~\cite{mot17_ref}. 
  % *We reproduce C-BIoU and report the best result obtained.
  }
  \label{tab:sota_mot17_test}
  \vspace*{-0.2cm}
\end{table}

\begin{table}
  \centering
  {\small{
  \begin{tabular}{@{}lccc@{}}
    \toprule
    Method & HOTA & IDF1 & MOTA \\
    \midrule
    % ByteTrack~\cite{bt_ref}   & \textcolor{lightgray}{71.5} & \textcolor{lightgray}{94.6} & \textcolor{lightgray}{-}\\
    ByteTrack~\cite{bt_ref}       & 72.1 & 75.3 & 94.5 \\
    % DeepSORT~\cite{deepsort_ref} & 69.6 & 94.8 & -\\
    OC-SORT~\cite{ocsort_ref}     & 82.0 & 76.3 & \textbf{98.3} \\
    % C-BIoU~\cite{cbiou_ref}    & \textcolor{lightgray}{\textbf{89.2}} & \textcolor{lightgray}{\textbf{86.1}} & \textcolor{lightgray}{\textbf{99.4}} \\
    C-BIoU~\cite{cbiou_ref} *     & 72.7 & 76.4 & 95.4 \\
    % Ours (McByte)             & 82.9 & 77.1 & 96.6 \\
    McByte (ours)                 & \textbf{85.0} & \textbf{79.9} & 96.8 \\
   
    \bottomrule
  \end{tabular}
  }}
  \caption{Comparing McByte with state-of-the-art tracking-by-detection algorithms on SoccerNet-tracking 2022 test set~\cite{soccernet-tracking2022_ref}. 
  % *We reproduce C-BIoU and report the best result obtained.
  }
  \label{tab:sota_soccernet_test}
\end{table}

\begin{table}
  \centering
  {\small{
  \begin{tabular}{@{}lcccc@{}}
    \toprule
    Method & HOTA & MOTA & HOTA & MOTA \\
    \midrule
     & \multicolumn{2}{c}{Pedestrian} & \multicolumn{2}{c}{Car} \\
    \midrule
    ByteTrack~\cite{bt_ref}             & 54.3 & 63.7 & 47.3 & 34.9 \\
    PermaTr~\cite{permatr_ref}          & 47.4 & 65.1 & 78.0 & 91.3\\
    OC-SORT~\cite{ocsort_ref}           & 54.7 & 65.1 & 76.5 & 90.3\\
    StrongSORT++~\cite{strongsort_ref}  & 54.5 & 67.4 & 77.8 & 90.4 \\
    % Ours (McByte)                       & 55.0 & 65.9\\
    McByte (ours)               & \textbf{57.0} & \textbf{68.9} & \textbf{80.8} & \textbf{92.5}\\
    % \textcolor{gray}{Ours (McByte) + C + R}      & \textcolor{gray}{55.7} & & \textcolor{gray}{66.8}\\
    \bottomrule
  \end{tabular}
  }}
  \caption{Comparing McByte with state-of-the-art tracking-by-detection algorithms on KITTI-tracking test set~\cite{kitti_ref}. KITTI evaluation server does not provide IDF1 scores.
  % and at the same time provides result for tracking pedestrian and car objects.
  }
  \vspace*{-0.5cm}
  \label{tab:sota_kitti_test}
\end{table}

The results of each variant are listed in \cref{tab:ablation_mask_constraints_gradual}. In variant a1, where only the TP mask signal is used for association, we can see that despite HOTA increase, IDF1 decreases with respect to the baseline. It is caused by the fact that the mask use is uncontrolled and chaotic. With TP mask possibly providing incorrect results, the association cues can be misleading. If we perform the association either based only on TP mask or only on IoU (depending on the availability of the mask), as in variant a2, we might face an inconsistency of the cues between tracklets and detections from the same frame and the next frames. This might lead to performance degradation. However, when we use properly both cues fusing the available information (variant a3), we can observe significant performance gain. We explain it as the algorithm is initially designed to work on bounding boxes while TP mask is a valuable guiding cue which can improve existing association mechanisms. When the mask signal is the only cue or not properly fused with the IoU cue, a lower performance might be obtained (as in a1 and a2). 
%
% The results for each variant are presented in \cref{tab:ablation_mask_constraints_gradual}. In variant a1, where only the mask signal is used for association, we see an increase in HOTA but a decrease in IDF1 compared to the baseline. This happens because the uncontrolled mask use can lead to incorrect associations, making the process chaotic. In variant a2, where the association switches between mask and IoU based on mask availability, the inconsistency between cues can cause mismatches within the same frame or across frames, leading to performance degradation. However, in variant a3, where both cues are properly fused, we observe a significant performance improvement. The algorithm, designed primarily for bounding boxes, effectively uses the mask as a guiding signal, improving association without overriding IoU. When the mask is used alone or without proper fusion (as in a1 and a2), it results in lower overall performance.

Adding the conditional check based on TP mask confidence (variant a4) further improves the performance, because sometimes mask might be uncertain or incorrect providing misleading association guidance. Adding the minimal $mm_{2}$ value check (variant a5) also provides performance gain, because this check filters out the tracklet TP masks which could be considered as a noise or tiny parts of people almost entirely occluded. Another performance gain can be observed with the minimal $mm_{1}$ value check (variant a6). This check determines if the TP mask of the tracked person is actually within the bounding box and not too much outside it. Since the detection box might not be perfect, we allow small parts of the tracklet mask to be outside the bounding box, but its major part must be within the bounding box, so that the TP mask can be used for guiding the association between the considered tracklet-detection pair. As it is shown, it further helps. Finally, we add the camera motion compensation (CMC), denoted as McByte in \cref{tab:ablation_mask_constraints_gradual}, which also provides some performance gain. 
We compare our McByte tracking algorithm to the baseline~\cite{bt_ref} across the four datasets, as shown in \cref{tab:ablation_mask_cmc}. Performance improvements are seen in all cases, though the gains vary by dataset due to different characteristics. For example, DanceTrack~\cite{dancetrack_ref} features non-linear motion, where the TP mask signal is particularly helpful in tracking occluded subjects, whereas IoU might struggle. In SoccerNet-tracking 2022~\cite{soccernet-tracking2022_ref}, where occlusions are fewer but motion is more abrupt (e.g., players running), the mask aids in successfully capturing players, despite similar outfits among team members.

% Less performance gain is observed on MOT17 than on the two previous datasets. We explain it that the scenes in MOT17 are more crowded and more people appear small on the scene, e.g. being far away. This makes it more difficult for the mask to capture them. The quality of detections, including small and uncertain bounding boxes, makes it more difficult for the mask to comply with our conditions ($mm_{1}$ and $mm_{2}$), which among the others, are based on the bounding boxes of detections. Lower performance gain can also be observed on KITTI-tracking. Reduced number of people (pedestrians) is present in this dataset compared to the others. In \cref{tab:sota_kitti_test}, we can observe that in case of the cars, which are more frequent in the KITTI-tracking datasets, higher performance gain can be obtained. Detection quality is lower than in the other datasets, and people appear usually smaller, while cars are bigger and easier to detect, explaining the improvement differences. The gain can be smaller or bigger on different datasets, however in all cases it is an improvement and not a degradation, which demonstrates the effectiveness and generality of our approach.
%
The performance gain on MOT17~\cite{mot17_ref} is smaller than on the previous datasets due to crowded scenes and smaller, distant people, making it harder for the TP mask to capture them accurately. The quality of detections, especially for small or uncertain bounding boxes, makes it difficult for the mask to comply with our conditions ($mm_{1}$ and $mm_{2}$), which rely on bounding boxes. KITTI-tracking shows a similar trend yet with higher gain, with less distant pedestrians and different quality of detections. 
% but larger, more easily detectable cars. As shown in \cref{tab:sota_kitti_test}, better detection quality for cars leads to higher performance gains. 
Although gains vary by dataset, McByte consistently improves performance (\cref{tab:ablation_mask_cmc}), demonstrating its robustness and general applicability. TP mask incorporated as an association cue helps to handle challenging scene situations such as high occlusions and reduced subject visibility as shown in \cref{fig:power_of_cutie}.

MyByte also performs better than the baseline when the public detections 
% of a different quality 
are given. Due to the space limits, we show it in \cref{sec:pub_dets}.

\subsection{Comparison with state of the art tracking-by-detection methods}
\label{sec:exps_sota_tr_by_det}

% We compare our McByte with state of the art tracking-by-detection algorithms on the test sets of all four datasets mentioned. The results are listed in \cref{tab:sota_dancetrack_test,tab:sota_mot17_test,tab:sota_soccernet_test,tab:sota_kitti_test}. Our proposed McByte reaches the highest HOTA and IDF1 scores on DanceTrack~\cite{dancetrack_ref} (\cref{tab:sota_dancetrack_test}), on SoccerNet-tracking 2022~\cite{soccernet-tracking2022_ref} (\cref{tab:sota_soccernet_test}) and the highest HOTA scores on KITTI-tracking~\cite{kitti_ref} (\cref{tab:sota_kitti_test}). 
%
We compare McByte with state-of-the-art tracking-by-detection algorithms across the test sets of the four diversed datasets. The results, listed in \cref{tab:sota_dancetrack_test,tab:sota_mot17_test,tab:sota_soccernet_test,tab:sota_kitti_test} show that McByte achieves the highest HOTA, IDF1 and MOTA scores on DanceTrack~\cite{dancetrack_ref} (\cref{tab:sota_dancetrack_test}), MOT17~\cite{mot17_ref} (\cref{tab:sota_mot17_test}) and KITTI-tracking~\cite{kitti_ref} (\cref{tab:sota_kitti_test}). In case of SoccerNet-tracking 2022~\cite{soccernet-tracking2022_ref} (\cref{tab:sota_soccernet_test}), McByte achieves the highest HOTA and IDF1 scores, and the second highest MOTA score. 
% As oracle detections, meaning perfect detections, are provided by the dataset authors~\cite{soccernet-tracking2022_ref}, we argue that MOTA is not very descriptive in terms of the actual tracking performance.

% KITTI-tracking test evaluation server does not provide IDF1 scores, while it evaluates the performance on two classes, pedestrian and car. We also run our baseline (ByteTrack~\cite{bt_ref}) on this dataset with the same detection set as for McByte and other methods to obtain the test result. The results of ByteTrack on the car class might appear unreasonably low. This algorithm is designed for tracking people and optimized around its pre-trained detections. When PermaTr~\cite{permatr_ref} detections, which include car detections, are inserted into ByteTrack, then it cannot generalize and handle them, resulting in low performance. The same detections induce comparable results with other methods on the pedestrian class.
%
The KITTI-tracking test server evaluates both pedestrian and car classes. We also test ByteTrack~\cite{bt_ref}, our baseline, with the same detection set for comparison. 
% ByteTrack performs poorly on the car class because it is designed for tracking people and is optimized for its pre-trained detections. 
ByteTrack performs poorly on the car class because its parameters are specifically set for tracking people.
% and the algorithm is optimized for its pre-trained detections.
When using PermaTr~\cite{permatr_ref} detections, which include cars, ByteTrack struggles to generalize, leading to lower performance, though it performs comparably on the pedestrian class. Our method performs well on pedestrians, but also on other objects, such as cars, which we demonstrate in \cref{tab:sota_kitti_test}.

% On MOT17~\cite{mot17_ref} test set, on the contrary to the basline~\cite{bt_ref} and the derived methods~\cite{strongsort_ref, ocsort_ref, deepocsort_ref, hybridsort_ref}, we do not tune the parameters per sequence. With our mask-based association cue, we reach improvement over baseline and comparable performance with other methods despite not tuning the parameters per sequence. Since the number of evaluations per user provided on the MOT17 test evaluation server~\cite{motchallenge_paper_ref} is highly limited, we are unable to evaluate all tracking-by-detection methods without their parameters tuned.
%
On the MOT17~\cite{mot17_ref} test set, unlike the baseline~\cite{bt_ref} and derived methods~\cite{strongsort_ref, ocsort_ref, deepocsort_ref, hybridsort_ref}, we do not tune parameters per sequence. We aim at developing a generalizable tracker without the need for changes on different sequences or datasets. We reach the best scores among the non-tuned trackers. We use the result of ByteTrack~\cite{bt_ref} not being tuned per sequence as reported in \cite{sushi_ref}. 
% (a global optimization method, see supplementary material, Appendix B). 
Juxtaposed with the methods which do tune their parameters per sequence, we achieve improvements over the baseline~\cite{bt_ref} and comparable performance with other methods. 
% Due to limited evaluations allowed on the MOT17 test server~\cite{motchallenge_paper_ref}, we are unable to evaluate all listed tracking-by-detection methods without sequence-specific tuning.

% We also remark that C-BIoU~\cite{cbiou_ref} method is placed with asterix in \cref{tab:sota_dancetrack_test,tab:sota_mot17_test,tab:sota_soccernet_test}. For this algorithm, no implementation is published and not all the necessary details for reproduction are provided. We reproduce the method based on all the description provided and report the best results we have managed to obtain. We describe it more in detail in the supplementary material, Appendix C.
%
Additionally, C-BIoU~\cite{cbiou_ref} is marked with an asterisk in \cref{tab:sota_dancetrack_test,tab:sota_mot17_test,tab:sota_soccernet_test}. As no implementation is published, and not all necessary details for reproduction are provided, we reproduce the method based on the available descriptions and report the best results we have obtained. We describe it more in detail in \cref{sec:cbiou_more_in_detail}.

\subsection{Comparison with other methods using mask}
\label{sec:exps_other_mask_methods}

We evaluate existing mask-based tracking methods: DEVA~\cite{deva_ref}, Grounded SAM 2~\cite{sam2_ref,gr_dino_ref} and MASA~\cite{masa_ref} on the MOT17 validation set. Each method is tested with its original settings and with YOLOX~\cite{yolox_ref} trained on MOT17~\cite{mot17_ref} from our baseline~\cite{bt_ref}, as used in McByte. Results in \cref{tab:deva_grsam2_masa_mot17val} show that all methods perform significantly worse than McByte.

DEVA~\cite{deva_ref} lacks a tracklet management system, leading to chaotic tracklet handling, resulting in negative MOTA scores, which occur when errors exceed the number of objects~\cite{motchallenge_paper_ref}. Adding YOLOX detections worsens the performance by introducing more redundant and noisy tracklets and masks.

Grounded SAM 2~\cite{sam2_ref,gr_dino_ref} struggles with segment-based tracking, where merging segments into full tracklets is unreliable. Though YOLOX detections improve the performance, the results remain 
% unsatisfactory and 
far below McByte's performance.

MASA~\cite{masa_ref} fails to handle longer occlusions and misses many detections in its original setup, yielding low MOT performance. While YOLOX detections improve its results, it still lags behind McByte and even the baseline due to ongoing challenges with occlusions and missed detections.

These results highlight that existing mask-based methods are unsuitable for MOT. In contrast, McByte effectively combines temporally propagated mask-based association with bounding box processing and tracklet management, making it better suited for MOT tasks.

More results and details on DEVA, Grounded SAM 2, and MASA experiments are available in \cref{sec:mask_based_systems_more_exps}.

% \begin{table}
%   \centering
%   {\small{
%   \begin{tabular}{@{}lccc@{}}
%     \toprule
%     Method & HOTA & MOTA & IDF1 \\
%     \midrule
%     DEVA, original settings        & 31.8 & -89.4 & 31.3 \\
%     DEVA, with YOLOX          & 24.7 & -239.7 & 20.4 \\
%     \midrule
%     Grounded SAM 2, original settings & 43.4 & 18.4 & 47.6 \\
%     % step=20
%     Grounded SAM 2, with YOLOX & 47.5 & 43.0 & 54.1 \\
%     % step=20
%     \midrule
%     MASA, original settings & 45.5 & 36.9 & 53.6 \\
%     MASA, with YOLOX & 63.5 & 74.0 & 73.6 \\
%     \midrule
%     McByte (ours)           & \textbf{69.9} & \textbf{78.5} & \textbf{82.8} \\
%     \bottomrule
%   \end{tabular}
%   }}
%   \caption{Comparison with the other tracking methods using segmentation mask: DEVA~\cite{deva_ref}, Grounded SAM 2~\cite{sam_ref,gr_dino_ref} and MASA~\cite{masa_ref} on MOT17 validation set~\cite{mot17_ref}. }
%   \label{tab:deva_grsam2_masa_mot17val}
% \end{table}
\begin{table}
  \centering
  {\small{
  \begin{tabular}{@{}lccc@{}}
    \toprule
    Method & HOTA & IDF1 & MOTA \\
    \midrule
    DEVA, original settings        & 31.8 & 31.3 & -89.4 \\
    DEVA, with YOLOX               & 24.7 & 20.4 & -239.7 \\
    \midrule
    Grounded SAM 2, original settings & 43.4 & 47.6 & 18.4 \\
    % step=20
    Grounded SAM 2, with YOLOX     & 47.5 & 54.1 & 43.0 \\
    % step=20
    \midrule
    MASA, original settings        & 45.5 & 53.6 & 36.9 \\
    MASA, with YOLOX               & 63.5 & 73.6 & 74.0 \\
    \midrule
    McByte (ours)                  & \textbf{69.9} & \textbf{82.8} & \textbf{78.5} \\
    \bottomrule
  \end{tabular}
  }}
  \caption{Comparison with the other tracking methods using segmentation mask: DEVA~\cite{deva_ref}, Grounded SAM 2~\cite{sam_ref,gr_dino_ref} and MASA~\cite{masa_ref} on MOT17 validation set~\cite{mot17_ref}. }
  \label{tab:deva_grsam2_masa_mot17val}
  \vspace*{-0.30cm}
\end{table}

\section{Conclusion}
\label{sec:conclusion}

% In this paper, we study the impact of a temporally propagated segmentation mask in the problem of MOT. We design a new mechanism to incorporate the mask as an association cue within a tracking-by-detection approach and fuse the mask information together with the bounding box information to enhance the tracking performance. The results conducted on four datasets demonstrate the gain of using the mask as a cue and show the boosted performance of our novel tracking algorithm incorporating it. Furthermore, our algorithm significantly outperforms on MOT the other tracking systems, which use the segmentation masks as the main guide or in less controllable manner. This suggests that the object segmentation mask, when used properly and carefully, can be a strong association cue for MOT. 
%
In this paper, we explore the use of a temporally propagated segmentation mask as an association cue for MOT. We develop a new mechanism that incorporates the mask propagation into a tracking-by-detection approach, fusing temporally propagated mask and bounding box information to improve bounding box-based multi-object tracking performance. Results from four datasets show the effectiveness of this approach, with our algorithm boosting MOT performance. Additionally, it outperforms other systems that rely primarily on segmentation masks, highlighting that when carefully managed, the mask can serve as a powerful association cue for MOT.

\section*{Acknowledgement}
This work was granted access to the HPC resources of IDRIS under the allocation 2024-AD011014370 made by GENCI. The work was performed within the 3IA Côte d'Azur funding.

% This work was performed using HPC resources from GENCI–IDRIS (Grant 2024-AD011014370).

% \newpage
% \pagebreak

%%%%%%%%% REFERENCES
{
    \small
    \bibliographystyle{ieeenat_fullname}
    \bibliography{main}
}

\newpage

\appendix
\begin{appendices}

This supplementary material contains the following appendices as referred in the main paper:

\begin{itemize}
  \item ~\ref{sec:mask_based_systems_more_exps} More experiments and details with mask-based tracking systems
  \item ~\ref{sec:sota_transformers_sushi} State-of-the-art comparison with transformer-based and other types of method
  \item ~\ref{sec:cbiou_more_in_detail} More information on C-BIoU
  \begin{itemize}
        \item ~\ref{sec:cbiou_more_in_detail_first_subsec} Our C-BIoU implementation and its performance
        \item ~\ref{sec:cbiou_more_in_detail_second_subsec} C-BIoU with temporally propagated mask as an association cue
    \end{itemize}
  % \item ~\ref{sec:param_tuning} Parameter tuning and its impacts
  \item ~\ref{sec:pub_dets} McByte and baseline on public detections
  \item ~\ref{sec:components_in_detail} McByte components more in detail
  \begin{itemize}
        \item ~\ref{sec:app_f_mask_management} Mask processing and management
        \item ~\ref{sec:app_f_mask_visib} Tracklet mask visibility at the scene
        \item ~\ref{sec:app_f_mask_conf} Mask confidence
        \item ~\ref{sec:app_f_mm1_mm2} Bounding box coverage (mm1) and mask fill ratio (mm2)
        \item ~\ref{sec:app_f_cmc} Camera motion compensation
        \item ~\ref{sec:app_f_more_visuals} Additional visual example
    \end{itemize}
\end{itemize}

\section{More experiments and details with mask-based tracking systems}
\label{sec:mask_based_systems_more_exps}

\begin{table*}
  \centering
  {\small{
  \begin{tabular}{@{}lccc@{}}
    \toprule
    Details & HOTA & IDF1 & MOTA \\
    \midrule
    \multicolumn{4}{c}{DEVA} \\
    \midrule
    GDino "person", th. 0.35 $\ddagger$ & \textbf{31.8} & \textbf{31.3} & \textbf{-89.4} \\
    YOLOX ByteTrack, th. 0.6 $\ddagger$ & 24.7 & 20.4 & -239.7 \\
    YOLOX ByteTrack, th. 0.7 & 27.0 & 23.7 & -187.8 \\
    \midrule
    \multicolumn{4}{c}{Grounded SAM 2} \\
    \midrule
    GDino "person", th. 0.25, step 20 $\ddagger$ & 43.4 & 47.6 & 18.4 \\
    GDino "person", th. 0.25, step 100  & 44.0 & 49.0 & 15.5 \\
    YOLOX ByteTrack, th. 0.25, step 20 & 46.4 & 51.6 & 36.0 \\
    YOLOX ByteTrack, th. 0.6, step 20 $\ddagger$ & \textbf{47.5} & 54.1 & 43.0 \\
    YOLOX ByteTrack, th. 0.7, step 20 & 47.4 & 54.1 & 44.3 \\
    YOLOX ByteTrack, th. 0.25, step 100 & 46.8 & 54.2 & 30.2 \\
    YOLOX ByteTrack, th. 0.6, step 100 & 47.4 & \textbf{54.9} & 34.8 \\
    YOLOX ByteTrack, th. 0.7, step 100 & 47.4 & \textbf{54.9} & 35.9 \\
    YOLOX ByteTrack, th. 0.25, step 1 & 43.0 & 43.9 & 36.2 \\
    YOLOX ByteTrack, th. 0.6, step 1 & 44.4 & 46.5 & 44.9 \\
    YOLOX ByteTrack, th. 0.7, step 1 & 44.3 & 46.7 & \textbf{46.5} \\
    \midrule
    \multicolumn{4}{c}{MASA} \\
    \midrule
    GDino feat. Detic-SwinB "person", th 0.2 & 46.8 & 52.1 & 24.3 \\
    GDino feat. YOLOX COCO, th 0.3 & 45.4 & 53.1 & 36.9 \\
    GDino feat. YOLOX ByteTrack, th 0.3 & 61.8 & 70.8 & 71.3 \\
    GDino feat. YOLOX ByteTrack, th 0.6  & 63.4 & 73.3 & 73.8 \\
    GDino feat. YOLOX ByteTrack, th 0.7  & 62.5 & 71.9 & 72.9 \\
    R50 feat. YOLOX COCO, th 0.3 $\ddagger$ & 45.5 & 53.6 & 36.9 \\
    R50 feat. YOLOX ByteTrack, th 0.3 & 62.5 & 72.0 & 71.5 \\
    R50 feat. YOLOX ByteTrack, th 0.6 $\ddagger$ & \textbf{63.5} & \textbf{73.6} & \textbf{74.0} \\
    R50 feat. YOLOX ByteTrack, th 0.7 & 62.6 & 72.3 & 73.0 \\
    \midrule
    \multicolumn{4}{c}{McByte} \\
    \midrule
    McByte (ours)           & \textbf{69.9} & \textbf{82.8} & \textbf{78.5} \\
    \bottomrule
  \end{tabular}
  }}
    \caption{Extended comparison with the other tracking methods using segmentation mask: DEVA~\cite{deva_ref}, Grounded SAM 2~\cite{sam_ref,gr_dino_ref} and MASA~\cite{masa_ref} on MOT17 validation set~\cite{mot17_ref}, while changing their parameters. $\ddagger$ denotes the variants reported in the main paper and in ~\cref{tab:deva_grsam2_masa_dt_val}.}
  \label{tab:deva_grsam2_masa_mot17val_extended}
\end{table*}

% \begin{table}
%   \centering
%   {\small{
%   \begin{tabular}{@{}lccc@{}}
%     \toprule
%     Method & HOTA & MOTA & IDF1 \\
%     \midrule
%     DEVA, original settings        & 21.9 & -347.1 & 15.8 \\
%     DEVA, with YOLOX          & 20.1 & -423.9 & 13.3 \\
%     \midrule
%     Grounded SAM 2, original settings & 51.3 & 73.5 & 48.0 \\
%     Grounded SAM 2, with YOLOX & 52.9 & 81.6 & 49.6 \\
%     \midrule
%     MASA, original settings & 38.2 & 71.9 & 34.9 \\
%     MASA, with YOLOX & 46.0 & 85.6 & 41.1 \\
%     \midrule
%     McByte (ours)          & \textbf{62.3} & \textbf{89.8} & \textbf{64.0} \\
%     \bottomrule
%   \end{tabular}
%   }}
%   \caption{Comparison with the other tracking methods using segmentation mask: DEVA~\cite{deva_ref}, Grounded SAM 2~\cite{sam_ref,gr_dino_ref} and MASA~\cite{masa_ref} on DanceTrack validation set~\cite{dancetrack_ref}. The reported variants correspond to the variants with $\ddagger$ symbol in ~\cref{tab:deva_grsam2_masa_mot17val_extended}}
%   \label{tab:deva_grsam2_masa_dt_val}
% \end{table}
\begin{table}
  \centering
  {\small{
  \begin{tabular}{@{}lccc@{}}
    \toprule
    Method & HOTA & IDF1 & MOTA \\
    \midrule
    DEVA, original settings        & 21.9 & 15.8 & -347.1 \\
    DEVA, with YOLOX               & 20.1 & 13.3 & -423.9 \\
    \midrule
    Grounded SAM 2, original settings & 51.3 & 48.0 & 73.5 \\
    Grounded SAM 2, with YOLOX        & 52.9 & 49.6 & 81.6 \\
    \midrule
    MASA, original settings           & 38.2 & 34.9 & 71.9 \\
    MASA, with YOLOX                  & 46.0 & 41.1 & 85.6 \\
    \midrule
    McByte (ours)                     & \textbf{62.3} & \textbf{64.0} & \textbf{89.8} \\
    \bottomrule
  \end{tabular}
  }}
  \caption{Comparison with the other tracking methods using segmentation mask: DEVA~\cite{deva_ref}, Grounded SAM 2~\cite{sam_ref,gr_dino_ref} and MASA~\cite{masa_ref} on DanceTrack validation set~\cite{dancetrack_ref}. The reported variants correspond to the variants with $\ddagger$ symbol in ~\cref{tab:deva_grsam2_masa_mot17val_extended}}
  \vspace*{-0.4cm}
  \label{tab:deva_grsam2_masa_dt_val}
\end{table}

%%% / %%%

% We evaluate DEVA~\cite{deva_ref}, Grounded SAM 2~\cite{sam2_ref, gr_dino_ref} and MASA~\cite{masa_ref} on MOT datasets. For every outputted bounding box on each frame, we save its data to a text file following the MOT format convention~\cite{motchallenge_paper_ref}. While Grounded SAM 2 and MASA provide IDs of the tracked objects, in case of DEVA, we use immutable and unique IDs of the propagated masks associated with each bounding box. 
%
We evaluate DEVA~\cite{deva_ref}, Grounded SAM 2~\cite{sam2_ref, gr_dino_ref}, and MASA~\cite{masa_ref} on MOT datasets, saving each bounding box output per frame in MOT format~\cite{motchallenge_paper_ref}. 
% Grounded SAM 2 and MASA provide object IDs, while for DEVA, we use unique, immutable IDs from the propagated masks associated with each bounding box.
The limitations of these methods when applied to MOT are described in \cref{sec:related_work__seg_mask_models,sec:exps_other_mask_methods} of the main paper.

% We provide more experiments to demonstrate careful exploration of the mask-based tracking systems and their performance differences with respect to our McByte. We provide more variants on MOT17~\cite{mot17_ref} validation set and an analogous table as in the main paper, yet on DanceTrack~\cite{dancetrack_ref} validation set.
%
We conduct additional experiments to thoroughly explore the performance differences between the mask-based tracking systems and our McByte. These include several variants on the MOT17~\cite{mot17_ref} validation set, as well as a corresponding table for the DanceTrack~\cite{dancetrack_ref} validation set, analogous to the one presented in the main paper.

% \cref{tab:deva_grsam2_masa_mot17val_extended} lists different variants run on MOT17 validation set, where we change the detectors and used parameters. $\ddagger$ denotes the variants that are included in the main paper.
%
\cref{tab:deva_grsam2_masa_mot17val_extended} presents various experimental variants on the MOT17 validation set, where different detectors and parameters are used. The variants marked with $\ddagger$ correspond to those discussed in the main paper.

% For DEVA, we first run it with the default settings, i.e. with the Grounding Dino~\cite{gr_dino_ref} detector with prompt "person". The confidence threshold for not suppressing the bounding boxes is 0.35. Next, we insert the YOLOX~\cite{yolox_ref} detector with the weights trained on the MOT17 dataset, provided by our baseline~\cite{bt_ref}. We run variants with two threshold: 0.6 and 0.7, which correspond to the values of high confidence detection threshold and new tracklet initialization in our method respectively.
%
For DEVA, we first run the default settings using the Grounding Dino~\cite{gr_dino_ref} detector with the "person" prompt and a confidence threshold of 0.35 to accept bounding boxes. Then, we replace it with the YOLOX~\cite{yolox_ref} detector, trained on the MOT17 dataset from our baseline~\cite{bt_ref}. We test two threshold values, 0.6 and 0.7. In our baseline, initialization of the new tracklets happens for the values 0.1 higher than the high confidence detection threshold. As we consider the default value of 0.6 for the latter (\cref{sec:exps_impl_details} in the main paper), we also experiment with the value of 0.7 with DEVA and other mask based systems.
% which correspond to the high-confidence detection threshold \textbf{\textcolor{red}{[REMOVE IT, for consistency with the main paper] and the new tracklet initialization}} in our method, respectively (Sec. 4.1 in the main paper).

% For Grounded SAM 2~\cite{sam2_ref,gr_dino_ref}, we consider the version of "Video Object Tracking with Continues ID" as specified on its github page\footnote{\url{https://github.com/IDEA-Research/Grounded-SAM-2}}. We first run the variant with original settings: using Grounding Dino~\cite{gr_dino_ref} detector with prompt "person", original confidence detection threshold 0.25 and the step value of 20. Step determines how often the detections are processed (e.g. every 20th frame) and objects are established to create the mask tracklets. It is considered as the segment length (recall tracking objects in segments mentioned in the main paper). Next, we run the analogous variant with step value of 100. 
%
For Grounded SAM 2~\cite{sam2_ref,gr_dino_ref}, we use the "Video Object Tracking with Continuous ID" version as specified on its GitHub page\footnote{\url{https://github.com/IDEA-Research/Grounded-SAM-2}}. Initially, we run it with the original settings, using the Grounding Dino~\cite{gr_dino_ref} detector with the "person" prompt, a confidence detection threshold of 0.25, and a step value of 20. The step value defines how often detections are processed (e.g., every 20th frame) to create mask tracklets, functioning as the segment length (we refer to tracking objects in segments mentioned in the main paper, \cref{sec:related_work__seg_mask_models}). We then test an analogous variant with a step value of 100.

% Further, we insert the YOLOX detectors with the weights from our baseline~\cite{bt_ref} and run variants with the step values of 20, 100 and 1 with different bounding box allowance thresholds 0.25 as well as 0.6 and 0.7 (as mentioned for DEVA). We also attempt to run the variant with the segment lenght of the whole video sequence, however it cannot be finished due to the extensive GPU memory requirement. Besides, it would only track the objects appearing during the first frame.
%
Next, we integrate YOLOX detector with weights from our baseline~\cite{bt_ref} and run variants with step values of 20, 100, and 1 (thus processing detections every frame), using different bounding box allowance thresholds of 0.25, 0.6, and 0.7 (analogous to the DEVA experiments). We also attempt to run a variant with the segment length set to the entire video sequence, but it fails due to excessive GPU memory requirements. Additionally, this setup would only track objects visible in the first frame.

% MASA~\cite{masa_ref} provides a few models for inference. We run variants with two different feature backbones: GroundingDINO~\cite{gr_dino_ref} (GDino) and ResNet-50~\cite{resnet_ref} (R50). We run the GroundingDINO variant with Detic-SwinB detector~\cite{detic_ref,swin_transformer_ref} with prompt "person" with the original detection confidence threshold of 0.2. We run an analogous variant, but with YOLOX detector trained on COCO~\cite{coco_dataset_ref} dataset as provided by the authors, with original threshold of 0.3.
%
MASA~\cite{masa_ref} offers several models for inference. We test variants using two different feature backbones: GroundingDINO~\cite{gr_dino_ref} (GDino) and ResNet-50~\cite{resnet_ref} (R50). For the GroundingDINO variant, we use the Detic-SwinB detector~\cite{detic_ref,swin_transformer_ref} with the "person" prompt, applying the original detection confidence threshold of 0.2. We also run a similar variant with the YOLOX detector trained on the COCO~\cite{coco_dataset_ref} dataset, as provided by the authors, using a confidence threshold of 0.3 default for this variant.

% Further, we insert YOLOX detector with the weights from our baseline~\cite{bt_ref} and run variants with different detection confidence thresholds: 0.3, 0.6 and 0.7 (as mentioned for DEVA and Grounded SAM 2). We also run the ResNet-50 feature variants with YOLOX COCO version (threshold 0.3) and weights as pre-trained by our baseline (thresholds 0.3, 0.6, 0.7).
%
Further, we incorporate the YOLOX detector with weights from our baseline~\cite{bt_ref} and test variants with detection confidence thresholds of 0.3, 0.6, and 0.7, analogously to DEVA and Grounded SAM 2. Additionally, we run the ResNet-50 feature variants with the YOLOX COCO model (threshold 0.3) and the baseline-pre-trained weights (thresholds 0.3, 0.6, 0.7).

% As it is demonstrated in \cref{tab:deva_grsam2_masa_mot17val_extended}, McByte performs better than the referenced mask-based systems, which makes it more suitable for MOT.
%
As shown in \cref{tab:deva_grsam2_masa_mot17val_extended}, McByte outperforms the referenced mask-based systems, making it more suitable for MOT.

% \cref{tab:deva_grsam2_masa_dt_val} lists the performance of DEVA, Grounded SAM 2 and MASA on DanceTrack~\cite{dancetrack_ref} validation set. The listed variants correspond to the ones with $\ddagger$ symbol in ~\cref{tab:deva_grsam2_masa_mot17val_extended} and thus to the variants reported in the main paper.
%
\cref{tab:deva_grsam2_masa_dt_val} presents the performance of DEVA, Grounded SAM 2, and MASA on the DanceTrack~\cite{dancetrack_ref} validation set. The listed variants correspond to those marked with $\ddagger$ in ~\cref{tab:deva_grsam2_masa_mot17val_extended} and are the ones reported in the main paper on MOT17.

% Also on DanceTrack, McByte manifests considerably higher performance demonstrating effectiveness and suitability for MOT.
% 
On DanceTrack, McByte also demonstrates significantly higher performance, reinforcing its effectiveness and suitability for MOT.

\section{State-of-the-art comparison with transformer-based and other types of method}
\label{sec:sota_transformers_sushi}

\begin{table}
  \centering
  {\small{
  \begin{tabular}{@{}lccc@{}}
    \toprule
    Method & HOTA & IDF1 & MOTA \\
    \midrule
    \multicolumn{4}{c}{Transformer-based} \\
    \midrule
    MOTR~\cite{motr_ref}                & 57.8 & 68.6 & 73.4 \\
    MeMOTR~\cite{memotr_ref}            & 58.8 & 71.5 & 72.8 \\
    MOTRv2~\cite{motrv2_ref}            & 62.0 & 75.0 & 78.6 \\
    MOTIP~\cite{motip_ref}              & 59.2 & 71.2 & 75.5 \\
    \midrule
    \multicolumn{4}{c}{Global optimization} \\
    \midrule
    SUSHI~\cite{sushi_ref}              & 66.5 & 83.1 & 81.1 \\
    \midrule
    \multicolumn{4}{c}{Joint detection and tracking} \\
    \midrule
    FairMOT~\cite{fairmot_ref}          & 59.3 & 72.3 & 73.7 \\
    RelationTrack~\cite{relationtrack_ref}          & 61.0 & 75.8 & 75.6 \\
    CenterTrack~\cite{centertrack_ref}          & 52.2 & 64.7 & 67.8 \\
    \midrule
    \multicolumn{4}{c}{Tracking-by-detection } \\
    \multicolumn{4}{c}{with parameter tuning per sequence} \\
    \midrule
    \textcolor{gray}{ByteTrack~\cite{bt_ref}}            & \textcolor{gray}{63.1} & \textcolor{gray}{77.3} & \textcolor{gray}{80.3} \\
    \textcolor{gray}{StrongSORT++~\cite{strongsort_ref}} & \textcolor{gray}{64.4} & \textcolor{gray}{79.5} & \textcolor{gray}{79.6} \\
    \textcolor{gray}{OC-SORT~\cite{ocsort_ref}}          & \textcolor{gray}{63.2} & \textcolor{gray}{77.5} & \textcolor{gray}{78.0} \\
    \textcolor{gray}{Deep OC-SORT~\cite{deepocsort_ref}} & \textcolor{gray}{64.9} & \textcolor{gray}{80.6} & \textcolor{gray}{79.4} \\
    \textcolor{gray}{Hybrid-SORT~\cite{hybridsort_ref}}  & \textcolor{gray}{64.0} & \textcolor{gray}{78.7} & \textcolor{gray}{79.9} \\
    \midrule
    \multicolumn{4}{c}{Tracking-by-detection } \\
    \multicolumn{4}{c}{without parameter tuning per sequence} \\
    \midrule
    ByteTrack~\cite{sushi_ref} & 62.8 & 77.1 & 78.9 \\
    % C-BIoU~\cite{cbiou_ref} & 64.1 & 79.7 & 81.1 \\
    C-BIoU~\cite{cbiou_ref} * & 62.4 & 77.1 & 79.5 \\
    McByte (ours)             & 64.2 & 79.4 & 80.2 \\
    \bottomrule
  \end{tabular}
    }}
  \caption{Extended state-of-the-art method comparison on MOT17~\cite{mot17_ref} test set.}
  \label{tab:sota_mot17_test_extended}
\end{table}

\begin{table}
  \centering
  {\small{
  \begin{tabular}{@{}lccc@{}}
    \toprule
    Method & HOTA & IDF1 & MOTA \\
    \midrule
    \multicolumn{4}{c}{Transformer-based} \\
    \midrule
    MOTR~\cite{motr_ref}                & 54.2 & 51.5 & 79.7 \\
    MeMOTR~\cite{memotr_ref}            & 63.4 & 65.5 & 85.4 \\
    MOTRv2~\cite{motrv2_ref}            & 73.4 & 76.0 & 92.1 \\
    MOTIP~\cite{motip_ref}              & 67.5 & 72.2 & 90.3 \\
    \midrule
    \multicolumn{4}{c}{Global optimization} \\
    \midrule
    SUSHI~\cite{sushi_ref}              & 63.3 & 63.4 & 88.7 \\
    \midrule
    \multicolumn{4}{c}{Joint detection and tracking} \\
    \midrule
    FairMOT~\cite{fairmot_ref}          & 39.7 & 40.8 & 82.2 \\
    CenterTrack~\cite{centertrack_ref}          & 41.8 & 35.7 & 86.8 \\
    \midrule
    \multicolumn{4}{c}{Tracking-by-detection} \\
    \midrule
    ByteTrack~\cite{bt_ref}             & 47.7 & 53.9 & 89.6 \\
    OC-SORT~\cite{ocsort_ref}           & 55.1 & 54.9 & 92.2 \\
    Deep OC-SORT~\cite{deepocsort_ref}  & 61.3 & 61.5 & 92.3 \\
    % C-BIoU~\cite{cbiou_ref}             & \textcolor{lightgray}{60.6} & \textcolor{lightgray}{61.6} & \textcolor{lightgray}{91.6} \\
    C-BIoU~\cite{cbiou_ref} *           & 45.8 & 52.0 & 88.4 \\
    StrongSORT++~\cite{strongsort_ref}  & 55.6 & 55.2 & 91.1 \\
    Hybrid-SORT~\cite{hybridsort_ref}   & 65.7 & 67.4 & 91.8 \\
    McByte (ours)                       & 67.1 & 68.1 & 92.9 \\
    \bottomrule
  \end{tabular}
    }}
  \caption{Extended state-of-the-art method comparison on DanceTrack~\cite{dancetrack_ref} test set.}
  \label{tab:sota_dancetrack_test_extended}
\end{table}

%%% / %%%

% There are MOT methods outside the tracking-by-detection domain that perform better than ours on some benchmarks, 
There exist MOT methods outside the tracking-by-detection domain manifesting performance differences,
but usually these methods are not directly comparable, because they make certain hypotheses, e.g. global optimization on the whole video. At the same time, these methods might perform visibly worse on some benchmarks as we discuss below. On the contrary, we stress that McByte performs well on all the discussed benchmarks (\cref{sec:exps_abl_stud,sec:exps_sota_tr_by_det} of the main paper). McByte is a tracking-by-detection approach, which is the main focus of our work. For an additional reference, though, we also list performance of the transformer-based, global optimization, and joint detection and tracking methods. 

\cref{tab:sota_mot17_test_extended} shows extended state-of-the art comparison on MOT17~\cite{mot17_ref} test set. Note that analogously to the main paper, we also put the result of ByteTrack~\cite{bt_ref} not being tuned per sequence as reported in \cite{sushi_ref}. Transformer-based methods perform visibly lower than the tracking-by-detection methods (including ours) as they struggle with the subjects frequently entering and leaving the scene. In contrast, SUSHI~\cite{sushi_ref}, which is a powerful global optimization approach, reaches highly satisfying performance. However, it accesses all the video frames at same time while processing detections and associating the tracklets, which makes it impossible to run in online settings. This makes the performance not directly comparable and therefore, we do not include it in the main paper. Current state-of-the-art joint detection and tracking methods generally perform lower than the tracking-by-detection methods. In this paradigm, the detection and association step is performed jointly. In our method, we perform these two steps separately and focus on the association part.
% , thus the performance is not directly comparable.

\cref{tab:sota_dancetrack_test_extended} presents extended state-of-the-art comparison on DanceTrack~\cite{dancetrack_ref} test set. As in this dataset the subjects remain mostly at the scene, the transformer-based methods performance is more satisfying. The performance of transformer-based methods can be both higher~\cite{motrv2_ref,motip_ref} or lower~\cite{motr_ref,memotr_ref} compared to the the tracking-by-detection methods. For similar reasons, the global optimization method, SUSHI~\cite{sushi_ref} can also perform higher than the other tracking-by-detection methods on this dataset, or lower, e.g. when compared to our method. On this dataset, joint detection and tracking methods also manifest lower performance than the tracking-by-detection methods. As in the case of the MOT17 benchmark, the performance of the mentioned different method types is not directly comparable and thus not included in the main paper.

% \begin{figure*}
% \centering
% \begin{tabular}{cc}
% \includegraphics[height=4.5cm]{images/MOT17-04_train_BT_YOLOX_001050.jpg}&
% \includegraphics[height=4.5cm]{images/MOT17-04_FRCNN_001050.png}
% \\
% % Private detections - YOLOX model provided by baseline~\cite{bt_ref}&Public detections - FRCNN model provided by MOT17~\cite{mot17_ref}.
% Private detections & Public detections \\ YOLOX model provided by baseline~\cite{bt_ref}& FRCNN model provided by MOT17~\cite{mot17_ref}
% \end{tabular}
% % \vspace*{0.3cm}
% \caption{An example of detection quality difference. It can be seen that considerably more bounding boxes are missing in case of public detections, which negatively impacts the MOT performance. Input image data from~\cite{mot17_ref}, sequence MOT17-04, last frame (1050). Best seen in color.}
% \label{fig:private_vs_public_dets}
% \end{figure*}

\section{More information on C-BIoU}
\label{sec:cbiou_more_in_detail}

\subsection{Our C-BIoU implementation and its performance}
\label{sec:cbiou_more_in_detail_first_subsec}

% C-BIoU~\cite{cbiou_ref} paper does not include publicly available implementation. Further, the details which are necessary for the code reproduction are not provided. We implement the C-BIoU based on all the information provided and report the best obtained results on the datasets on which this algorithm has been evaluated. These results are denoted with an asterisk in the main paper and in the \cref{tab:sota_mot17_test,tab:sota_dancetrack_test} of this supplementary material.
% 
The C-BIoU~\cite{cbiou_ref} paper does not provide a public implementation or sufficient details for code reproduction. We implement C-BIoU based on the available information and report the best results on the datasets evaluated by the original algorithm. These results, marked with an asterisk, are shown in the main paper (\cref{tab:sota_dancetrack_test,tab:sota_mot17_test,tab:sota_soccernet_test}) and in \cref{tab:sota_mot17_test_extended,tab:sota_dancetrack_test_extended,tab:cbiou_with_mask} of this supplementary material.

Based on the descriptions in the C-BIoU work~\cite{cbiou_ref}, especially regarding tracklet management and tracklet-detection association, the core of the method strongly resembles ByteTrack~\cite{bt_ref}. Thus, we extend ByteTrack by incorporating cascaded buffered intersection over union~\cite{cbiou_ref} for the association process. Where C-BIoU does not specify parameters, we default to those used in ByteTrack. For the MOT17 test set, we do not tune detection confidence parameters per sequence (referred in \cref{sec:related_work__tr_by_det} of the main paper) and apply a fixed 0.6 threshold for all sequences and datasets. This is the default value of ByteTrack, i.e. when ByteTrack is run on datasets without tuning per sequence.
Regarding the two buffering scales in C-BIoU, we use the values 0.3 and 0.4, which were reported to result in the best HOTA on the DanceTrack validation set in the C-BIoU paper~\cite{cbiou_ref}. Although we also test the values 0.3 and 0.5 (as mentioned in the framework figure from C-BIoU), the 0.3 and 0.4 combination performs slightly better in our experiments.
Additionally, following C-BIoU’s supplementary material, we adopt the most optimal \textit{max\_age} values per dataset. The \textit{max\_age} parameter defines the maximum lifespan (in frames) of a tracklet that has not been matched to a detection before it is terminated and removed. For DanceTrack~\cite{dancetrack_ref}, we set \textit{max\_age} to 100, and for SoccerNet-tracking 2022~\cite{soccernet-tracking2022_ref}, we set it to 60. In case of MOT17~\cite{mot17_ref}, where this value is not directly specified, we test three different \textit{max\_age} values: 20, 60, and 100 (as listed in C-BIoU’s supplementary material). We select the value that results in the highest performance on the MOT17 validation set, which is 60, and then use this value for testing on the MOT17 test set to report the final results. 

% The results are listed in Tabs. 3 to 5 in the main paper. It can be seen that the with respect to the baseline, ByteTrack~\cite{bt_ref}, the improvement is not very high (Tab. 5 in the main paper) or slight degradation occurs (Tabs. 3 and 4 in the main paper). A possible explanation is the impact tuning (or its lack) of the parameters which have not been specified in the C-BIoU paper. More specificaly, with buffered intersection over union (IoU), the search space for the association is extended, more potential associations are considered and the IoU-based distances are lower. However, the matching thresholds for the linear association steps (the Hungarian Algorithm~\cite{hungarianalg_ref}), which exclude some of the potential yet unprobale matches are not specified. Using the default values of these thresholds from the baseline~\cite{bt_ref} results in allowing more matches and possibly more unresolved and misleading ambiguities. Hence, the performance might degrade.
% 
The results are presented in \cref{tab:sota_dancetrack_test,tab:sota_mot17_test,tab:sota_soccernet_test} of the main paper. Compared to the baseline, ByteTrack~\cite{bt_ref}, the improvement is modest (\cref{tab:sota_soccernet_test}) or shows slight degradation (\cref{tab:sota_dancetrack_test,tab:sota_mot17_test}). This could be due to tuning issues, as many parameters were not specified in the C-BIoU paper. Specifically, with buffered intersection over union (IoU), the association search space expands, leading to more potential associations and lower IoU-based distances. However, the matching thresholds for the linear association steps (using the Hungarian Algorithm~\cite{hungarianalg_ref}), which filter out unlikely matches, are not provided. Default thresholds from ByteTrack~\cite{bt_ref} may allow more matches, introducing unresolved ambiguities and potentially degrading performance.

We do not evaluate C-BIoU on KITTI-tracking test set, because the test evaluation server has a very limited number of submissions per person~\cite{kitti_ref} (less than MOT17 test server~\cite{motchallenge_paper_ref,mot17_ref}). We use these submissions for McByte and our baseline~\cite{bt_ref}. Besides, the results on KITTI-tracking test set are not reported in the work of C-BIoU~\cite{cbiou_ref}.

\subsection{C-BIoU with temporally propagated mask as an association cue}
\label{sec:cbiou_more_in_detail_second_subsec}

\begin{table}
  \centering
  {\small{
  \begin{tabular}{@{}lccc@{}}
    \toprule
    Method & HOTA & IDF1 & MOTA \\
    \midrule
    \multicolumn{4}{c}{DanceTrack test set} \\
    \midrule
    C-BIoU~\cite{cbiou_ref} *      & 45.8 & 52.0 & 88.4 \\
    C-BIoU~\cite{cbiou_ref} * with our TP mask cue        & \textbf{67.6} & \textbf{70.2} & \textbf{92.4} \\
    \midrule
    \multicolumn{4}{c}{SoccerNet-tracking 2022 test set} \\
    \midrule
    C-BIoU~\cite{cbiou_ref} *      & 72.7 & 76.4 & 95.4 \\
    C-BIoU~\cite{cbiou_ref} * with our TP mask cue        & \textbf{84.6} & \textbf{79.0} & \textbf{98.2} \\
    \bottomrule
  \end{tabular}
  }}
  \caption{Comaprison of another tracking algorithm~\cite{cbiou_ref}, with and without our approach of temporally propagated (TP) mask as an association cue (\cref{sec:cbiou_more_in_detail_second_subsec}).}
  \label{tab:cbiou_with_mask}
\end{table}

We also evaluate temporally propagated (TP) mask as an association cue within the C-BIoU~\cite{cbiou_ref} tracker to demonstrate the generalizability and advantage of our approach within MOT. We follow exactly the same steps as with incorporating the TP mask in our baseline~\cite{bt_ref} for McByte. ~\cref{tab:cbiou_with_mask} includes the performance comparison of the C-BIoU tracker with and without the TP mask as a cue. The results show that our TP mask-based approach is beneficial beyond single baseline and that it can improve MOT performance of other trackers as well.

As in the other result tables, the asterisk means that no official implementation is published and that we reproduce the C-BIoU method based on the available information, described in detail in~\cref{sec:cbiou_more_in_detail_first_subsec}. MOT17~\cite{mot17_ref} and KITTI-tracking~\cite{kitti_ref} test evaluation servers allow limited submissions which we use for McByte and our baseline~\cite{bt_ref}. Therefore, we report the results of this study on DanceTrack~\cite{dancetrack_ref} and SoccerNet-tracking 2022~\cite{soccernet-tracking2022_ref} test sets.

\begin{figure*}
\centering
\begin{tabular}{cc}
\includegraphics[height=4.5cm]{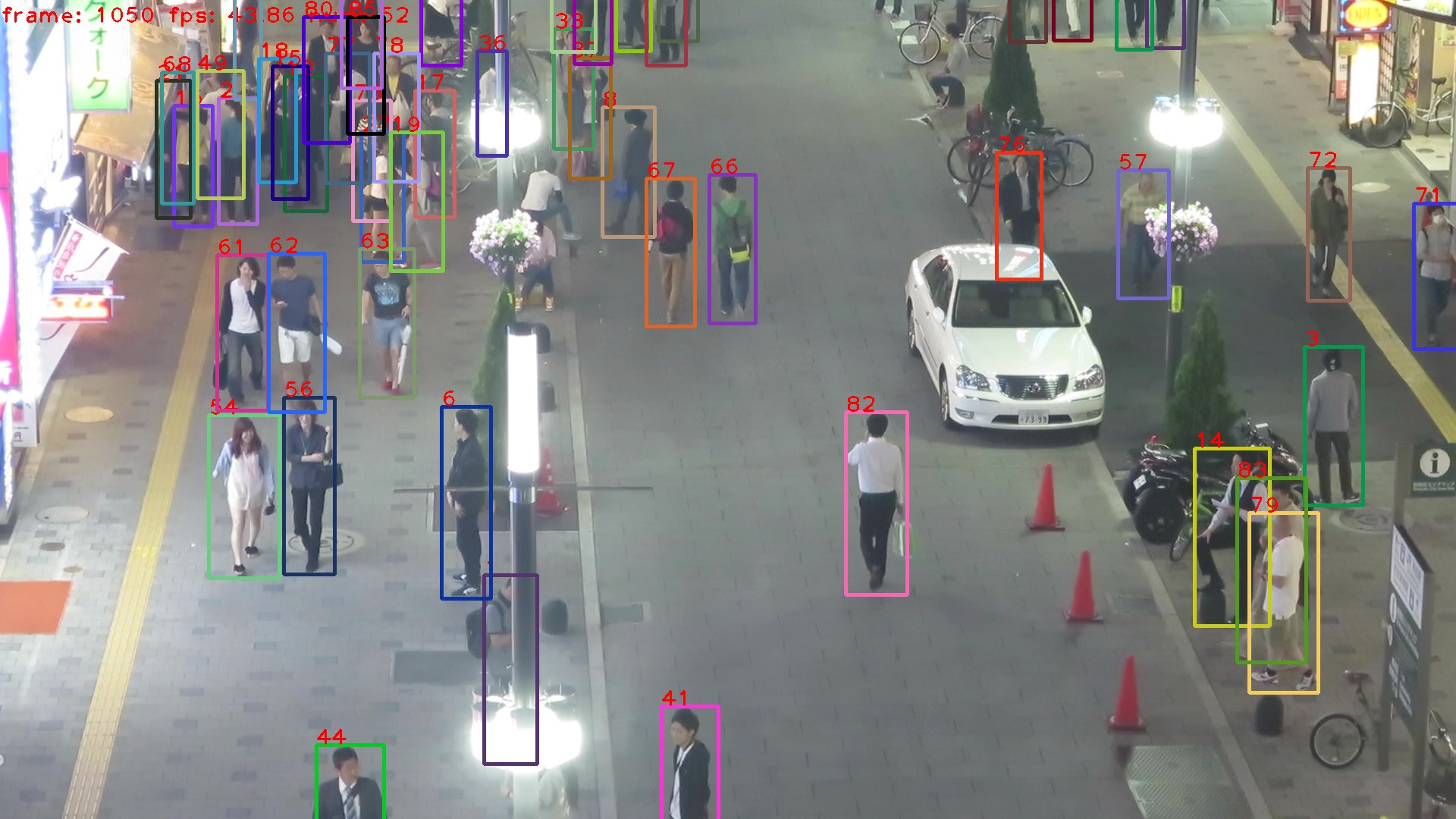}&
\includegraphics[height=4.5cm]{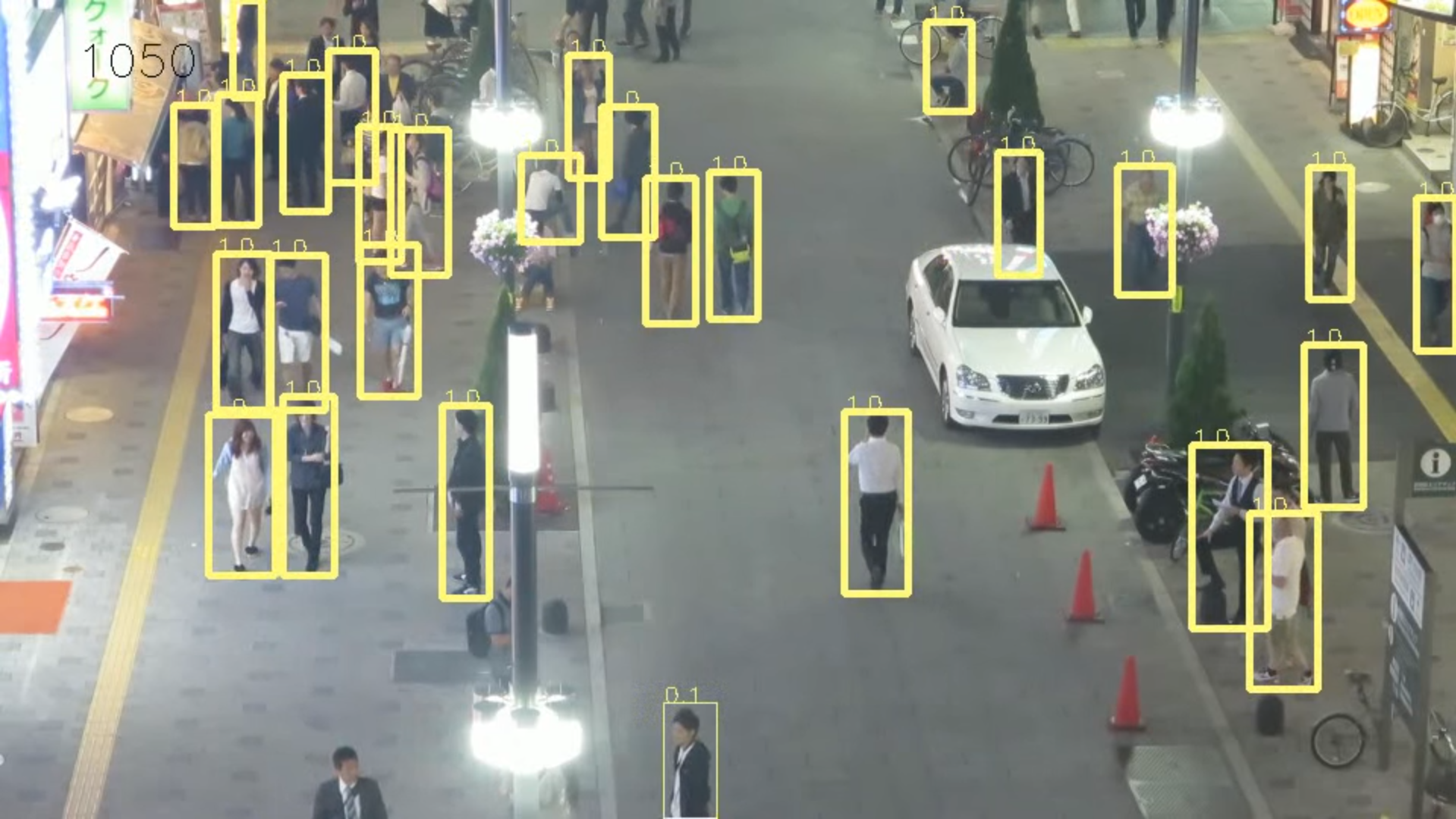}
\\
% Private detections - YOLOX model provided by baseline~\cite{bt_ref}&Public detections - FRCNN model provided by MOT17~\cite{mot17_ref}.
Private detections & Public detections \\ YOLOX model provided by baseline~\cite{bt_ref}& FRCNN model provided by MOT17~\cite{mot17_ref}
\end{tabular}
% \vspace*{0.3cm}
\caption{An example of detection quality difference. It can be seen that considerably more bounding boxes are missing in case of public detections, which negatively impacts the MOT performance. Input image data from~\cite{mot17_ref}, sequence MOT17-04, last frame (1050). Best seen in color.}
\label{fig:private_vs_public_dets}
\end{figure*}

\section{McByte and baseline on public detections}
\label{sec:pub_dets}

We also evaluate McByte and the baseline~\cite{bt_ref} on the MOT17~\cite{mot17_ref} validation set using the public detections provided by the dataset authors. The results are shown in \cref{tab:ablation_priv_pub_dets}, using the FRCNN model variant~\cite{frcnn_ref}. There are significant performance differences for both the baseline and McByte when using public detections, which are of lower quality compared to the private YOLOX~\cite{yolox_ref} detections pre-trained on MOT17 by the baseline~\cite{bt_ref}. Public detections tend to miss more objects, leading to many false negatives.
An example comparing the quality of private and public detections is illustrated in \cref{fig:private_vs_public_dets}. Since tracklets and their temporally propagated masks can only be initiated when a bounding box is detected, McByte’s relative performance gain over the baseline is slightly reduced when using public detections. However, even in this scenario, the temporally propagated mask-based association cue helps resolve ambiguities and improves overall tracking performance, as shown in \cref{tab:ablation_priv_pub_dets}.

% \begin{table}
%   \centering
%   {\small{
%   \begin{tabular}{@{}lccc@{}}
%     \toprule
%     Method & HOTA & MOTA & IDF1 \\
%     \midrule
%     \multicolumn{4}{c}{MOT17 val, private detections -  YOLOX~\cite{yolox_ref} from baseline~\cite{bt_ref}} \\
%     \midrule
%     baseline      & 68.4 & 78.2 & 80.2 \\
%     McByte & \textbf{69.9} & \textbf{78.5} & \textbf{82.8} \\
%     \midrule
%     \multicolumn{4}{c}{MOT17 val, public detections - FRCNN from~\cite{mot17_ref}} \\
%     \midrule
%     baseline      & 49.0 & 44.5 & 55.3 \\
%     McByte  & \textbf{50.1} & \textbf{44.6} & \textbf{56.0} \\
%     % \midrule
%     % \multicolumn{4}{c}{ \textbf{\textcolor{red}{[NEW!] }}MOT17 val, public detections - SDP} \\
%     % \midrule
%     % baseline      & 57.2 & 63.3 & 67.8 \\
%     % baseline + M  & 57.4 & 63.1 & 69.7 \\
%     \bottomrule
%   \end{tabular}
%   }}
%   \caption{ByteTrack and McByte with private and public detections on MOT17-val}
%   \label{tab:ablation_priv_pub_dets}
% \end{table}

\section{McByte components more in detail}
\label{sec:components_in_detail}

\begin{table}
  \centering
  {\small{
  \begin{tabular}{@{}lccc@{}}
    \toprule
    Method & HOTA & IDF1 & MOTA \\
    \midrule
    \multicolumn{4}{c}{MOT17 val, private detections -  YOLOX~\cite{yolox_ref} from baseline~\cite{bt_ref}} \\
    \midrule
   Baseline      & 68.4 & 80.2 & 78.2 \\
    McByte        & \textbf{69.9} & \textbf{82.8} & \textbf{78.5} \\
    \midrule
    \multicolumn{4}{c}{MOT17 val, public detections - FRCNN from~\cite{mot17_ref}} \\
    \midrule
    Baseline      & 49.0 & 55.3 & 44.5 \\
    McByte        & \textbf{50.1} & \textbf{56.0} & \textbf{44.6} \\
    % \midrule
    % \multicolumn{4}{c}{ \textbf{\textcolor{red}{[NEW!] }}MOT17 val, public detections - SDP} \\
    % \midrule
    % baseline      & 57.2 & 67.8 & 63.3 \\
    % baseline + M  & 57.4 & 69.7 & 63.1 \\
    \bottomrule
  \end{tabular}
  }}
  \caption{McByte and baseline~\cite{bt_ref} with private and public detections on MOT17-val}
  \label{tab:ablation_priv_pub_dets}
\end{table}

\subsection{Mask processing and management}
\label{sec:app_f_mask_management}

We provide more details on mask handling introduced in \cref{sec:method_mask_management} of the main paper. New objects might appear in any frame on the scene. It involves initiating new tracklets and necessitates creating new segmentation masks. We use an image segmentation model and provide the bounding boxes of new tracklets and the currently processed frame to create new masks. Further, a mask temporal propagator receives new initial masks and the frame at which they were created. Subsequently, the propagator receives the next frame, for which it returns the updated mask predictions, e.g. based on the object movement and scenery change. Each tracklet has an ID and each mask included within the mask temporal propagator also gets its ID. We link masks to the tracklets based on their IDs.

Tracklets might also be marked inactive and hence removed at any frame, e.g. when a tracklet has not been matched to any detection for a long time. It induces removing the tracklets and necessitates purging the masks from the mask temporal propagator. 
As masks are removed from the propagator, the IDs of the remaining masks change (shift) due to the memory allocation mechanisms~\cite{xmem_ref,cutie_ref}. 
% to cover the empty, removed slots. 
We update the links between tracklet IDs and mask IDs to consistently keep the same masks assigned to the tracklets. Tracklet IDs are immutable. 
For our experimentation, we use SAM~\cite{sam_ref} as the image segmentation model and Cutie~\cite{cutie_ref} as the mask temporal propagator.

\begin{figure*}
\centering
\includegraphics
[width=17.5cm]
{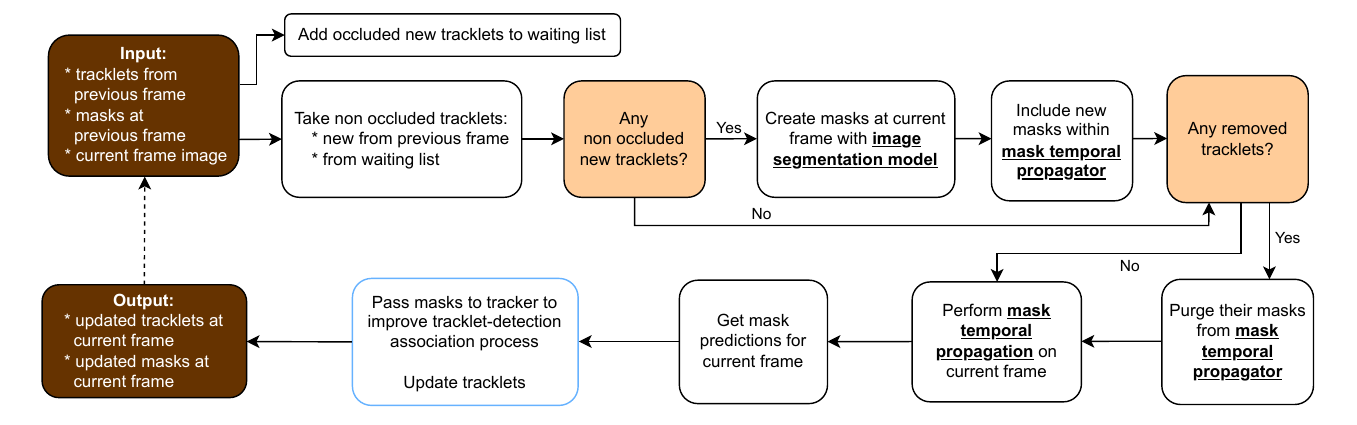}
% \vspace*{0.3cm}
\caption{Diagram illustrating the mask processing and management process as described in~\cref{sec:app_f_mask_management}. Tracklets influence the mask management process, while masks influence the tracklet association and updating. The process of temporally propagated masks influencing the association process (the blue part) is illustrated in \cref{fig:diagram_tracking_pipeline} (main paper).}
\label{fig:diagram_mask_management}
\end{figure*}

The mask handling process is illustrated in~\cref{fig:diagram_mask_management}. We process each frame of the video sequence separately. We check if there are any new initialized tracklets from the previously processed frame. If they are strongly occluded, their IDs are added to the waiting list. The tracklet positions change over the frames and they might become less occluded during the next frames. We take currently non-occluded tracklets, both newly initialized and from the waiting list, and create their masks with the image segmentation model. We include the new masks within the mask temporal propagator.

Further, we check if any tracklets have been removed at the previous frame and if so, we purge their masks from the mask temporal propagator. Regardless of adding or removing any masks, we perform mask temporal propagation to get the updated mask predictions for the current frame. The temporally propagated masks are passed to the tracker where they are used to improve the tracklet-detection association process. The process of temporally propagated masks influencing the association process is illustrated in \cref{fig:diagram_tracking_pipeline} (main paper) and described in \cref{sec:method_mask_use} (main paper). After the association, the tracker returns updated tracklets. Output tracklets and masks from the current frame processing constitute the input for the next frame processing. During the whole process, state of the tracklets influences the mask management, i.e. the mask creation and purging. At the same time, the masks influence the tracklet association and updating, improving the MOT performance.

\subsection{Tracklet mask visibility at the scene}
\label{sec:app_f_mask_visib}

% It might happen that the mask of a considered tracklet does not appear at the scene at the current frame. It might be the case when the subject is entirely or highly occluded or when the mask temporal propagator fails to predict the mask, e.g. for a very small visible part of the subject or because of difficult conditions such as dense crowds or very poor lighting. If the mask is not predicted for the tracklet, all its potential associations with detections are computed based on the intersection over union (IoU) as in the baseline~\cite{bt_ref}.
%
In order to ensure reliable use of the temporally propagated (\textbf{TP}) mask, we apply several conditions, which we mention in \cref{sec:method_mask_use} in the main paper. We describe them more in detail in the following subsections starting with tracklet mask visibility at the scene.

In some cases, a tracklet's TP mask may not appear in the current frame. This can happen if the subject is completely or mostly occluded, or if the mask temporal propagator fails to predict the mask due to challenges like small visible areas, dense crowds, or poor lighting conditions. When no mask is predicted for a tracklet, its associations with detections are calculated using intersection over union (IoU), as done in the baseline~\cite{bt_ref}.

\subsection{Mask confidence}
\label{sec:app_f_mask_conf}

% For each predicted pixel of the mask coming from the mask temporal propagator, one can obtain its confidence probability~\cite{xmem_ref,cutie_ref}. For each predicted mask of a tracklet, we average this value over all the pixels to determine if the mask prediction is confident enough and if it can be reliably used within the tracklet-detection association process. If this probability is below our defined threshold (0.6, see Sec. 4.1 in the main paper), then pure IoU is used for the association of the tracklet.
% 
Each predicted pixel in the mask provided by the mask temporal propagator is assigned a confidence probability~\cite{xmem_ref,cutie_ref}. To determine if the TP mask prediction is reliable enough to be used in the tracklet-detection association process, we average these confidence values across all pixels for each tracklet mask separately. 
% If the average confidence falls below our threshold \textcolor{red}{\textbf{[\underline{remove} or update after adding extra exps]}(set to 0.6, as explained in Sec. 4.1 of the main paper)}, the association relies solely on the intersection over union (IoU) score.
If the average confidence is too low, the association relies solely on the intersection over union (IoU) score.

\begin{figure}
\centering
\includegraphics
[width=8cm]
{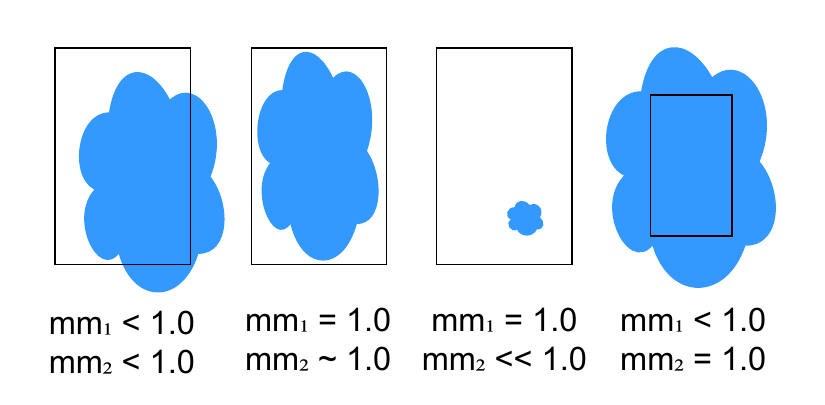}
% \vspace*{0.3cm}
\caption{Cases showing the differences in $mm_{1}$ and $mm_{2}$ values of a temporally propagated mask (in blue) within a bounding box (\cref{sec:app_f_mm1_mm2}). The most optimal case for the mask to provide a good guidance is the second one from the left, where
% when the mask is sufficiently big and when it is covered by a bounding box as much as possible, i.e 
both $mm_{1}$ and $mm_{2}$ are as close to $1$ as possible.}
% \vspace*{-0.5cm}
\label{fig:mm1_mm2_appendix}
\end{figure}

\begin{figure*}
\centering
\begin{tabular}{cccc}
\includegraphics[height=3.5cm]{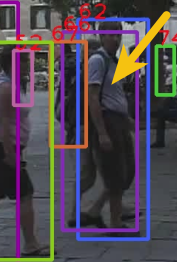}&
\includegraphics[height=3.5cm]{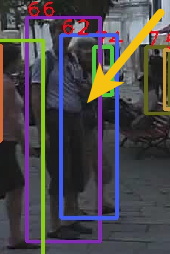}&
\includegraphics[height=3.5cm]{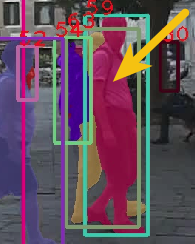}&
\includegraphics[height=3.5cm]{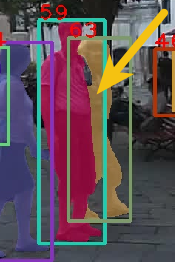}
\\
Frame 459 (baseline)&Frame 475 (baseline)&Frame 459 (McByte)&Frame 475 (McByte)
\end{tabular}
\vspace*{0.3cm}
\caption{Visual output comparison between the baseline and McByte. With the temporally propagated mask guidance, McByte can handle the association of an ambiguous set of bounding boxes 
% and reduce the identity switches
- see the subjects with IDs 59 and 63 on the output of McByte. Input image data from~\cite{mot17_ref}. Best seen in color.}
\label{fig:visual_differences_2}
\end{figure*}

\subsection{Bounding box coverage (mm1) and mask fill ratio (mm2)}
\label{sec:app_f_mm1_mm2}

% For describing $mm_{1}$ and $mm_{2}$ (Sec. 3.3 in the main paper) more in detail, we refer to the figure showing their different values. For the reader convenience, we place this figure also in the supplementary material, see \cref{fig:mm1_mm2}.
% 
For a more detailed explanation of $mm_{1}$ and $mm_{2}$ (see \cref{sec:method_mask_use} in the main paper), we refer to the accompanying figure that illustrates their varying values. For the reader's convenience, this figure is also included in the supplementary material, see \cref{fig:mm1_mm2_appendix}.

% In the main paper we define the bounding box coverage of the mask, referred to as $mm_{1}$. It is computed for each tracklet-detection pair (if the mask is present at the scene and is confident enough) and it determines what percentage of the given tracklet mask is present within the bounding box of the considered detection. If the mask is placed entirely within the bounding box, the $mm_{1}$ value will be 1.0, as presented in  \cref{fig:mm1_mm2}, second and third case from the left. In case the mask sticks out of the bounding box, $mm_{1}$ value will be lower, as in the first and fourth case from the left in \cref{fig:mm1_mm2}. Sometimes, bounding box might be a bit erroneous, thus we allow $mm_{1}$ to be slightly below 1.0 (we set the threshold as 0.9, see Sec. 4.1 in the main paper). However, the mask cannot stick out too much of the box, because then it might belong to a tracklet, which matches better another detection. If this is the case, the mask is not used to influence the cost of the match between the considered tracklet-detection pair.
%
In the main paper, we define the bounding box coverage of the mask, referred to as $mm_{1}$. It measures the percentage of the tracklet's TP mask that falls within the detection's bounding box for each tracklet-detection pair (if the mask is visible and confident). When the entire TP mask is within the detection’s bounding box, the $mm_{1}$ value is 1.0, as seen in the second and third case from the left in \cref{fig:mm1_mm2_appendix}. However, if part of the TP mask extends outside the bounding box, the $mm_{1}$ value decreases, as shown in the first and fourth case from the left in \cref{fig:mm1_mm2_appendix}. Bounding boxes can sometimes be slightly inaccurate, so we allow the $mm_{1}$ value to be slightly below 1.0.
% , \textcolor{red}{\textbf{[\underline{remove} or update after adding extra exps]}, with a threshold of 0.9 (see  Sec. 4.1 in the main paper)}. 
However, if too much of the TP mask extends beyond the bounding box, it could indicate that the mask belongs to a different detection, in which case the mask is excluded from influencing the cost of the particular tracklet-detection pair match.

% Further, in the main paper we define the mask fill ratio of the bounding box, referred to as $mm_{2}$. Analogously, it is computed for each tracklet-detection pair. It determines the percentage of the detection bounding box being covered by the mask. If the mask covers a small part of the bounding box, the value of $mm_{2}$ will be low, e.g. as in  \cref{fig:mm1_mm2} in the third case from the left. It might be a noisy or incorrect mask prediction or part of another mask being included within the considered bounding box. Therefore, we impose a condition expecting a minimal value for $mm_{2}$ (0.05, see Sec. 4.1 in the main paper) to prevent misleading guidance.
%
In the main paper, we also define the mask fill ratio of the bounding box, referred to as $mm_{2}$. This is calculated for each tracklet-detection pair and measures the percentage of the detection bounding box covered by the TP mask. If the TP mask only covers a small portion of the bounding box, the $mm_{2}$ value will be low, as illustrated in the third case from the left in \cref{fig:mm1_mm2_appendix}. This can indicate a noisy or incorrect mask prediction or that part of another TP mask is overlapping the bounding box. 
% \textcolor{red}{\textbf{[\underline{remove} or update after adding extra exps]}To avoid misleading associations, we set a minimum threshold for $mm_{2}$ (0.05, see  Sec. 4.1 in the main paper) to ensure reliable guidance.}
In our conditions (\cref{sec:method_mask_use} of the main paper) we do not allow the mask to influence the tracklet-detection association in such cases.

% For the higher coverage, the value of $mm_{2}$ will be naturally higher. However, this value is not sufficient on its own to determine if the mask tracklet matches well the considered detection bounding box. In \cref{fig:mm1_mm2}, in the first case from the left, the mask covers the major part of the bounding box, but it considerably sticks out of the box. In the fourth case from the left, the value of $mm_{2}$ is 1.0, meaning it covers the bounding box entirely, but it sticks out of it even more, thus likely not indicating a good match. Such cases may occur, when one subject is covered by another one, which at the moment is closer to the camera. $mm_{2}$ can be close to 1.0, but it is unlikely to be exactly 1.0 for good matches, since the considered subjects (people) are not rectangular, thus not covering their bounding boxes entirely. 
% 
For higher coverage, the $mm_{2}$ value naturally increases. However, this value alone is not enough to confirm whether the tracklet, to which the TP mask is linked, properly matches the detection bounding box. For example, in the first case from the left in \cref{fig:mm1_mm2_appendix}, the TP mask covers most of the bounding box, but it extends significantly beyond it. In the fourth case, $mm_{2}$ is 1.0, indicating that the TP mask fully covers the bounding box, but it sticks out even more, making it an unlikely good match. Such situations can occur when one subject is covered by another, closer to the camera. While $mm_{2}$ can approach 1.0, it is rare to see exactly 1.0 in good matches, as subjects (such as people) are not rectangular and do not fully align with their bounding boxes.

% For the best influence of the mask for a tracklet-detection pair, both $mm_{1}$ and $mm_{2}$ values should be as close to 1.0 as possible. An example is presented in  \cref{fig:mm1_mm2} in the second case from the left.
% 
For the TP mask to have the best influence on the tracklet-detection pair match, both $mm_{1}$ and $mm_{2}$ should be as close to 1.0 as possible, as shown in the second case from the left in \cref{fig:mm1_mm2_appendix}.

% If both $mm_{1}$ and $mm_{2}$ meet their value conditions, the cost matrix is updated as mentioned in the main paper (Sec. 3.3 in the main paper):
% \begin{equation}
%      costs^{i,j} = costs^{i,j} - mm^{i,j}_{2}
%      \label{eq:cost_matrix_update}
% \end{equation}
% Since as we described above, the high values of $mm_{1}$ might be misleading for the mask-based match between a tracklet and a detection, we use it only as a gating condition to prevent the matches based on the masks filling the bounding box and considerably sticking out of it (as in \cref{fig:mm1_mm2}, fourth case from the left). Instead, we use $mm_{2}$, for for which higher values denote higher match with a bounding box, as long as $mm_{1}$ is high enough as well.
% 
If both $mm_{1}$ and $mm_{2}$ meet the required conditions, the cost matrix is updated as described in the main paper (\cref{sec:method_mask_use}): 
\begin{equation}
     costs^{i,j} = costs^{i,j} - mm^{i,j}_{2}
     \label{eq:cost_matrix_update_appendix}
\end{equation}
Since high $mm_{1}$ values can be misleading when a TP mask significantly extends beyond the bounding box, $mm_{1}$ is used only as a gating condition to prevent poor matches (e.g., the fourth case from the left in \cref{fig:mm1_mm2_appendix}). Instead, we use $mm_{2}$, which better reflects the match quality between a tracklet TP mask and a detection bounding box, provided $mm_{1}$ is sufficiently high.

\subsection{Camera motion compensation}
\label{sec:app_f_cmc}

% For handling camera motion, we use camera motion compensation (CMC) as described in Sec. 3.4 in the main paper. Our approach follows the existing methods~\cite{strongsort_ref,deepocsort_ref}. Specifically, based on the camera movement, we compute the warp (transformation) matrix based on the extracted image features and apply this matrix to the predicted tracklet bounding boxes to take the camera movement into account. The tracklet predictions based on Kalman Filter~\cite{kf_ref} and associations with detections become more accurate which further improves the tracking performance. For the key-point extraction, we use the ORB (Oriented FAST and Rotated BRIEF) approach~\cite{orb_ref}. For more details we refer the reader to the related works~\cite{strongsort_ref, deepocsort_ref}. 
% 
For handling camera motion, we use camera motion compensation (CMC), as outlined in \cref{sec:method_mcbyte_with_cmc} of the main paper. Our approach follows the existing methods~\cite{strongsort_ref,deepocsort_ref}. Specifically, we compute a warp (transformation) matrix that accounts for camera movement, based on extracted image features, and apply this matrix to the predicted tracklet bounding boxes. This helps adjust for the camera motion, making the tracklet predictions from the Kalman Filter~\cite{kf_ref} and the associations with detections more accurate, improving the overall tracking performance. For key-point extraction, we use the ORB (Oriented FAST and Rotated BRIEF) approach~\cite{orb_ref}. For further details, we refer the reader to the related works~\cite{strongsort_ref, deepocsort_ref}.

\subsection{Additional visual example}
\label{sec:app_f_more_visuals}

% \begin{figure*}
% \centering
% \begin{tabular}{cccc}
% \includegraphics[height=4.0cm]{images/bt2a.png}&
% \includegraphics[height=4.0cm]{images/bt2b.png}&
% \includegraphics[height=4.0cm]{images/ours2a.png}&
% \includegraphics[height=4.0cm]{images/ours2b.png}
% \\
% Frame 459 (baseline)&Frame 475 (baseline)&Frame 459 (McByte)&Frame 475 (McByte)
% \end{tabular}
% \vspace*{0.3cm}
% \caption{Visual output comparison between the baseline and McByte. With the temporally propagated mask guidance, McByte can handle the association of an ambiguous set of bounding boxes 
% % and reduce the identity switches
% - see the subjects with IDs 59 and 63 on the output of McByte. Input image data from~\cite{mot17_ref}. Best seen in color.}
% \label{fig:visual_differences_2}
% \end{figure*}

In the main paper, we discuss that McByte can handle challenging scenarios due to the temporally propagated mask signal used in the controlled manner as an association cue (\cref{sec:method_mask_use}). \cref{fig:visual_differences_2} in this supplementary material illustrates our method handling association of ambiguous boxes, improving over the baseline. A figure illustrating handling long term occlusions in the crowd is placed in the main paper (\cref{sec:method_mask_use}).

\end{appendices}

% WARNING: do not forget to delete the supplementary pages from your submission 
% \input{sec/X_suppl}

\end{document}